\documentclass{article}

\usepackage{times}
\usepackage{graphicx} % more modern
\usepackage{subfigure}
\usepackage{verbatim} 

\usepackage[sort&compress]{natbib}

\usepackage{algorithm}
\usepackage{algorithmic}

\usepackage{hyperref}       % hyperlinks
\usepackage{url}            % simple URL typesetting
\usepackage{booktabs}       % professional-quality tables
\usepackage{amsfonts}       % blackboard math symbols
\usepackage{nicefrac}       % compact symbols for 1/2, etc.
\usepackage{microtype}      % microtypography

\usepackage{times}
\usepackage{calc}
\usepackage{mathtools}
\usepackage{color}
\usepackage{graphicx} % more modern
\usepackage{subfigure}
\usepackage{algorithm}
\usepackage{algorithmic}
\usepackage{booktabs}
\usepackage{enumitem}
\usepackage{bbm}
\usepackage{multirow}
\usepackage{pifont}
\usepackage{wrapfig}
\usepackage{longtable}
\usepackage{soul}
\usepackage{kky}
\usepackage{gittinsDefns}

\usepackage{hyperref}

\usepackage[accepted]{icml2017}

\icmltitlerunning{Bayesian Optimisation with Continuous Approximations}
\begin{document} 

\twocolumn[
\icmltitle{Multi-fidelity Bayesian Optimisation with Continuous Approximations}



% you can specify symbols, otherwise they are numbered in order
\icmlsetsymbol{equal}{*}

\begin{icmlauthorlist}
\icmlauthor{Kirthevasan Kandasamy}{cmu}
\icmlauthor{Gautam Dasarathy}{rice}
\icmlauthor{Jeff Schneider}{cmu}
\icmlauthor{Barnab\'as P\'oczos}{cmu}
% \icmlauthor{Eee Pppp}{ed}
\end{icmlauthorlist}

\icmlaffiliation{cmu}{Carnegie Mellon University, Pittsburgh PA, USA}
\icmlaffiliation{rice}{Rice University, Houston TX, USA}

\icmlcorrespondingauthor{Kirthevasan Kandasamy}{\incmtt{kandasamy@cmu.edu}}
% \icmlcorrespondingauthor{Eee Pppp}{ep@eden.co.uk}

\icmlkeywords{Multi-fidelity Optimisation, Bayesian Optimisation}

\vskip 0.3in
]



\printAffiliationsAndNotice{\icmlEqualContribution} % otherwise use the standard text.
%\footnotetext{hi}

\newcommand{\imarrwthree}{2.18in}
\newcommand{\imhspthree}{-0.05in}
\newcommand{\imleftspace}{-0.10in}
\newcommand{\imtextspace}{-0.15in}
\newcommand{\imsinglecol}{2.495in}
\newcommand{\imcaptionspace}{-0.1in}

\newcommand{\insertFigFidelSpace}{
\begin{figure}
\centering
\includegraphics[width=2.2in]{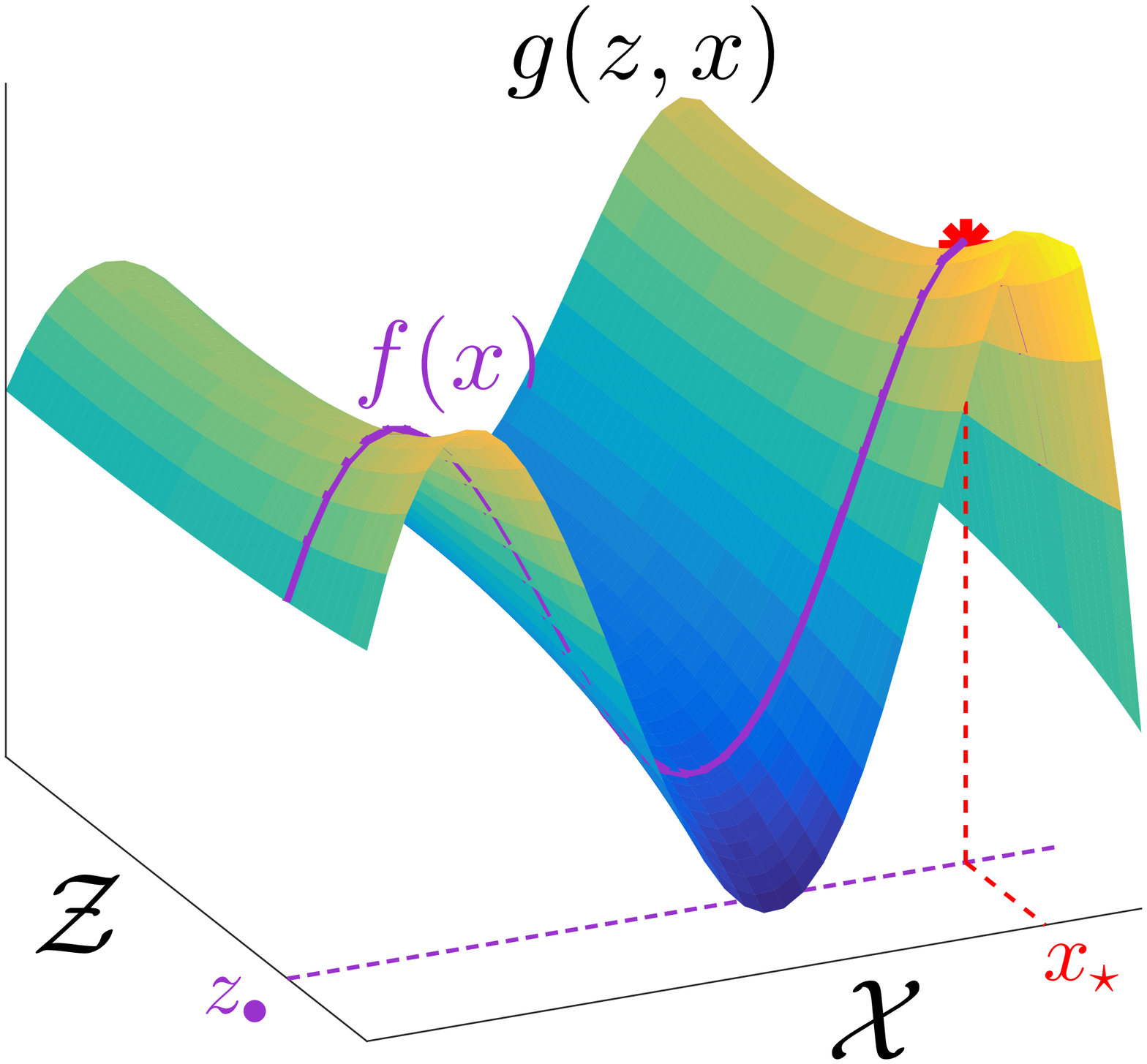}
\vspace{\imcaptionspace}
\caption{\small
\label{fig:fidelSpace}
$g:\Zcal\times\Xcal\rightarrow\RR$ is a function defined on the product space of the
fidelity space $\Zcal$ and domain $\Xcal$.
The purple line is $f(x) = g(\zhf, x)$.
We wish to find the maximiser $\xopt \in \argmax_{x\in\Xcal} f(x)$.
The multi-fidelity framework is attractive when $g$ is smooth across $\Zcal$ as
illustrated in the figure.
\vspace{\imtextspace}
}
\end{figure}
}

\newcommand{\insertFigBWSample}{
\begin{figure}
\centering
\includegraphics[width=2.0in]{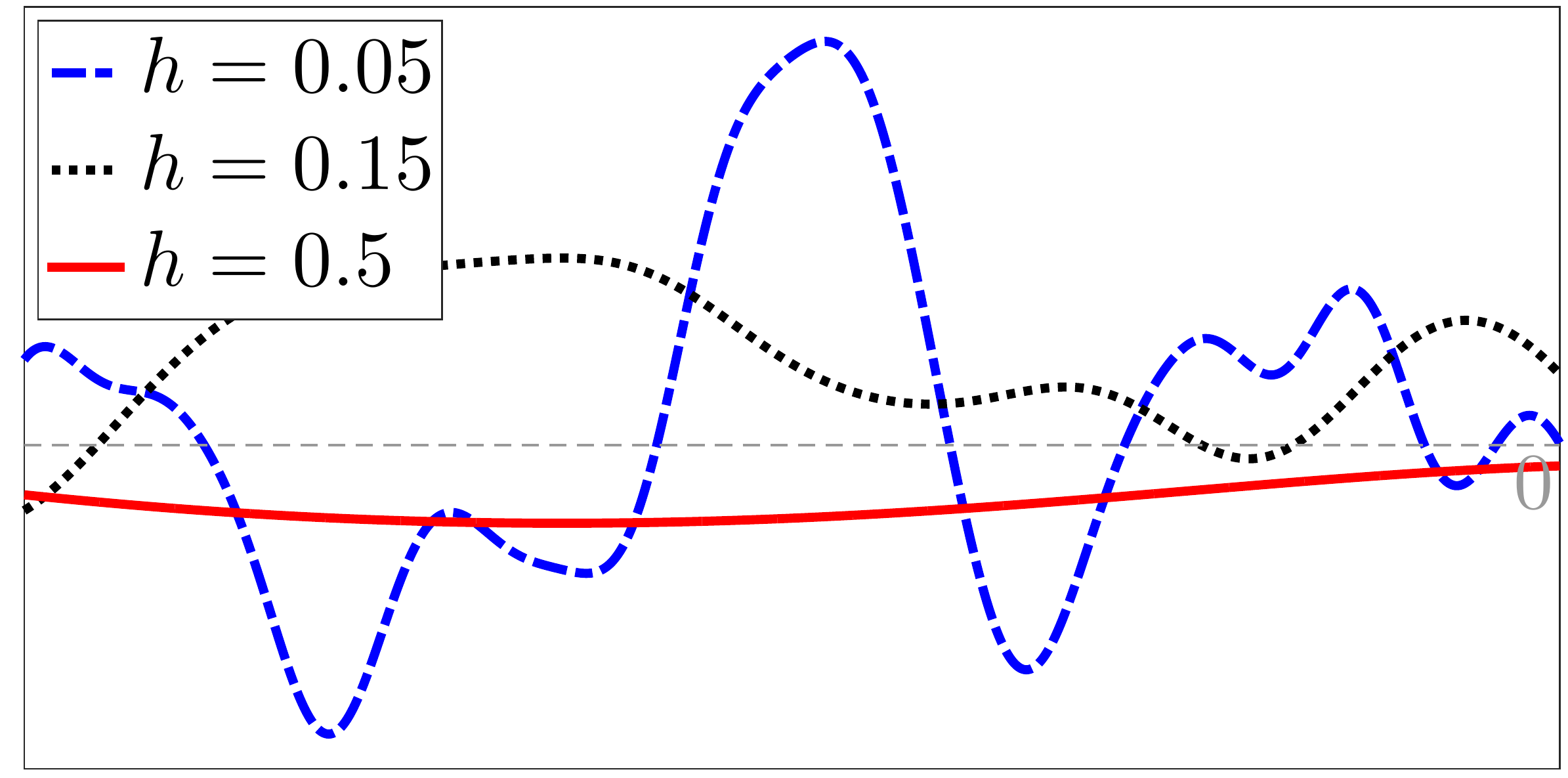}
\vspace{\imcaptionspace}
\caption{\small
\label{fig:gpBWSample}
Samples drawn from a GP with $0$ mean and SE kernel with bandwidths $h=0.01,h=0.15,0.5$.
Samples tend to be smoother across the domain for large bandwidths.
% \vspace{\imtextspace}
\vspace{-0.2in}
}
\end{figure}
}

\newcommand{\insertToyResultsFigure}{
\begin{figure*}
\centering
\hspace{\imleftspace}
\subfigure{
  \includegraphics[width=\imarrwthree]{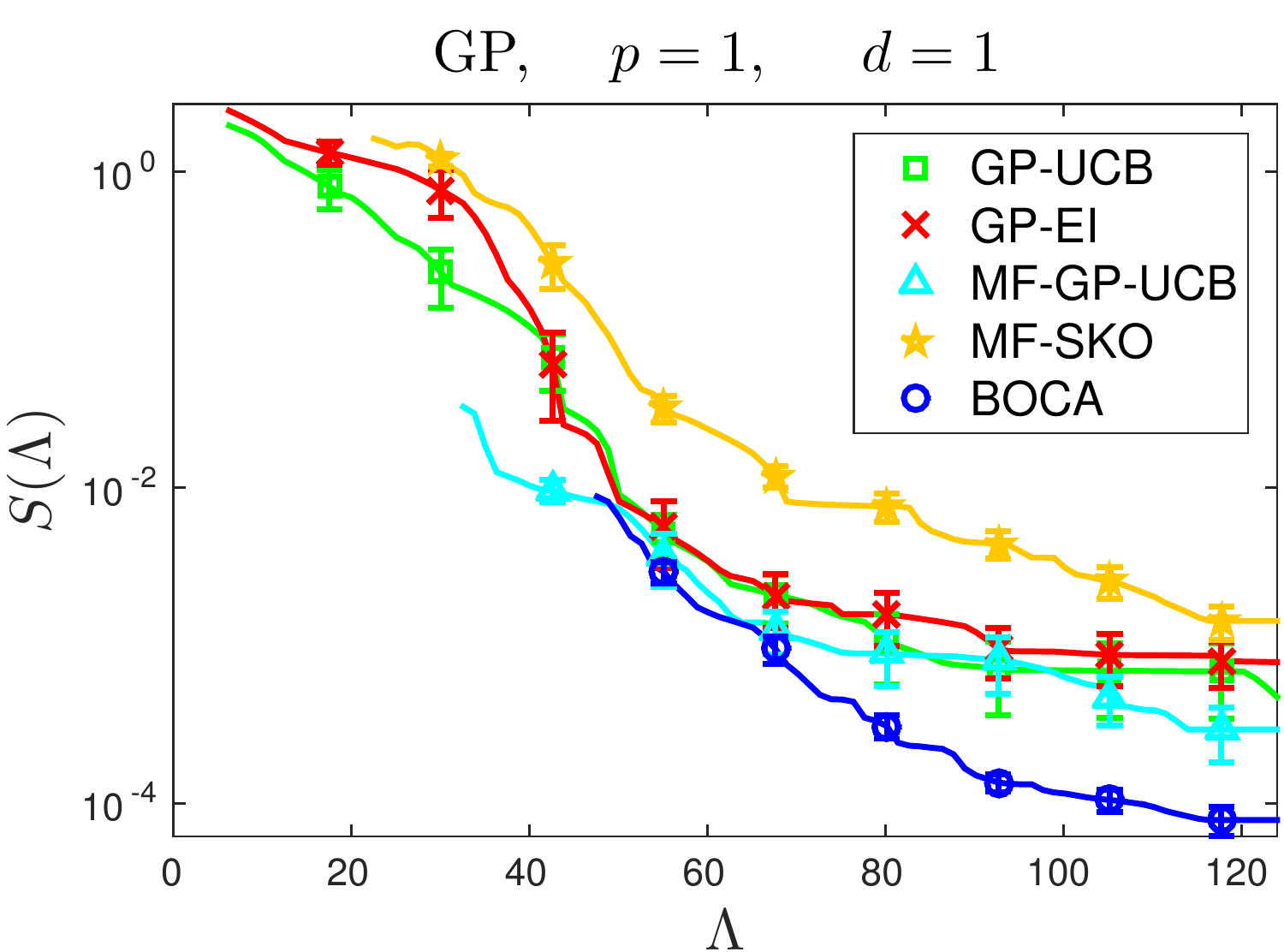} \hspace{\imhspthree}
}
\subfigure{
  \includegraphics[width=\imarrwthree]{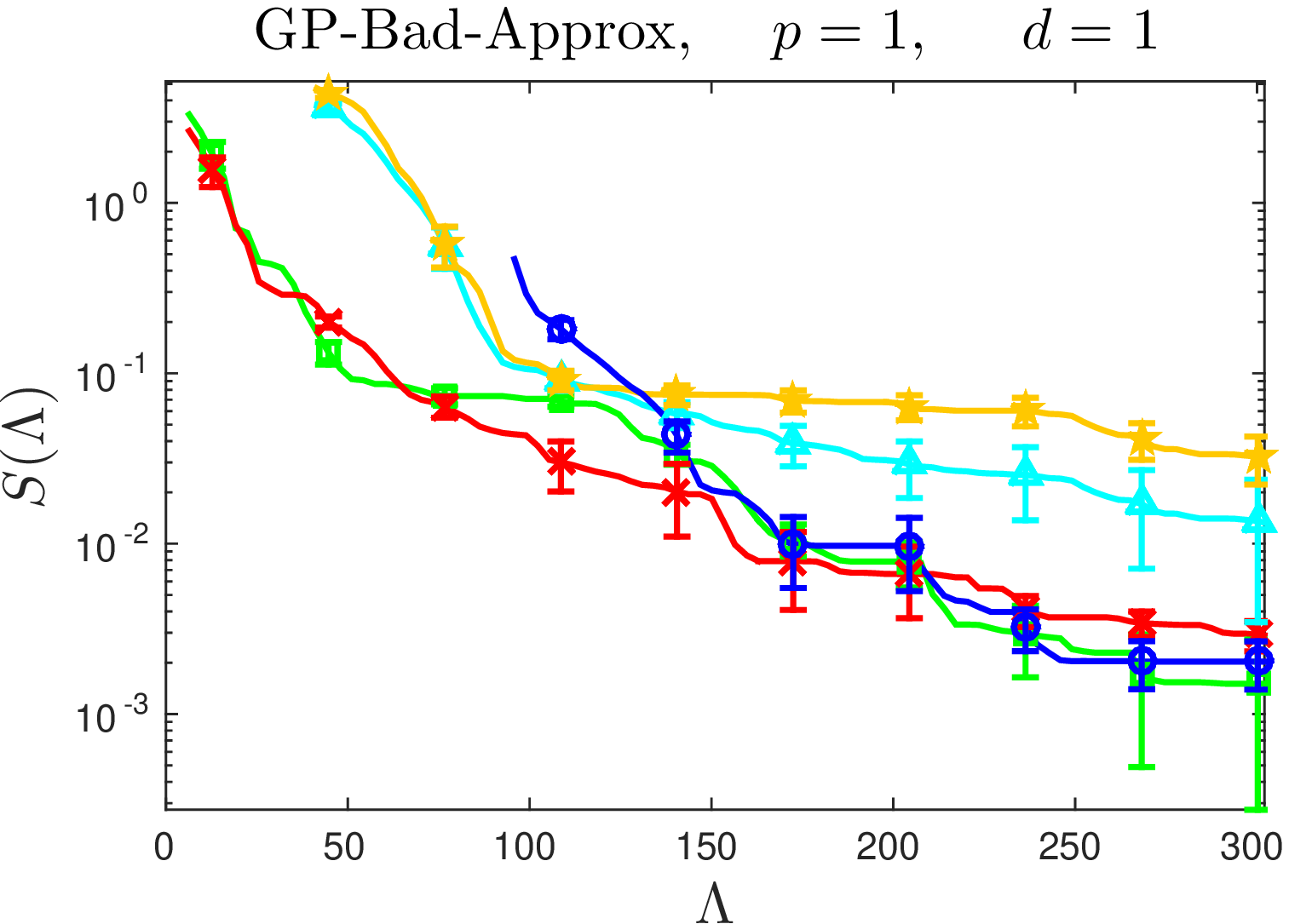} \hspace{\imhspthree}
}
\subfigure{
  \includegraphics[width=\imarrwthree]{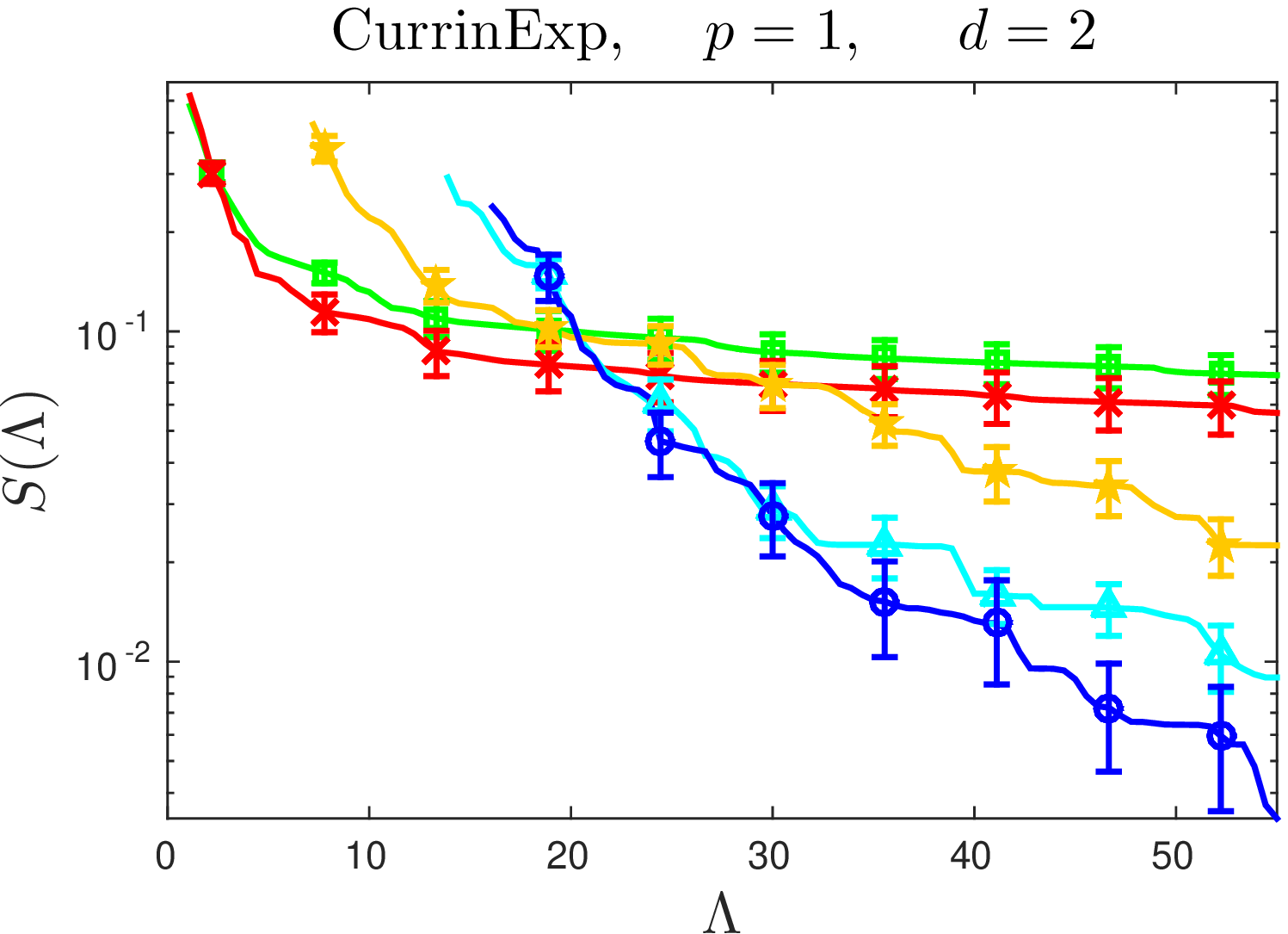} \hspace{\imhspthree}
} \\[\imcaptionspace]
\hspace{\imleftspace}
\subfigure{
  \includegraphics[width=\imarrwthree]{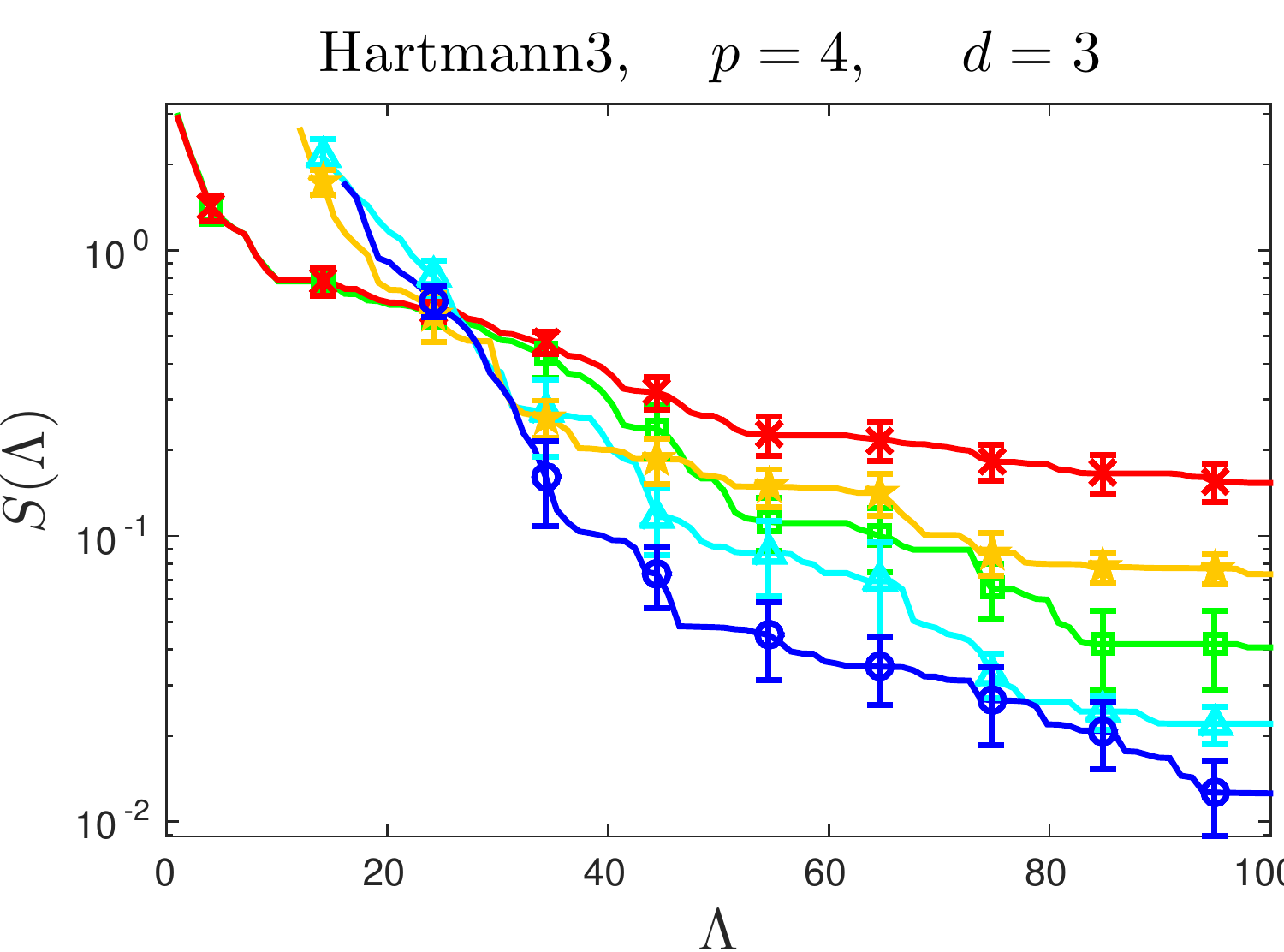} \hspace{\imhspthree}
}
\subfigure{
  \includegraphics[width=\imarrwthree]{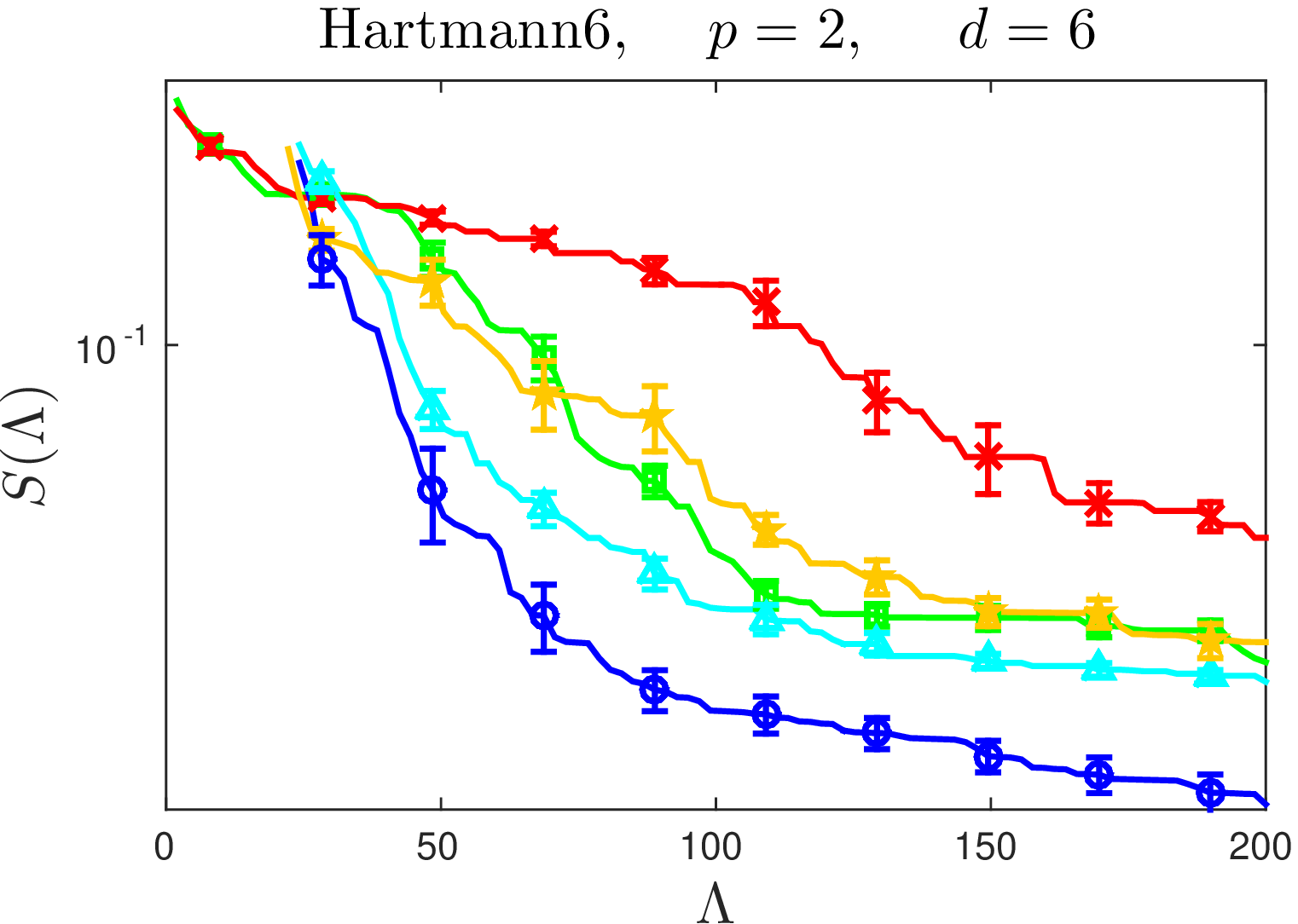} \hspace{\imhspthree}
}
\subfigure{
  \includegraphics[width=\imarrwthree]{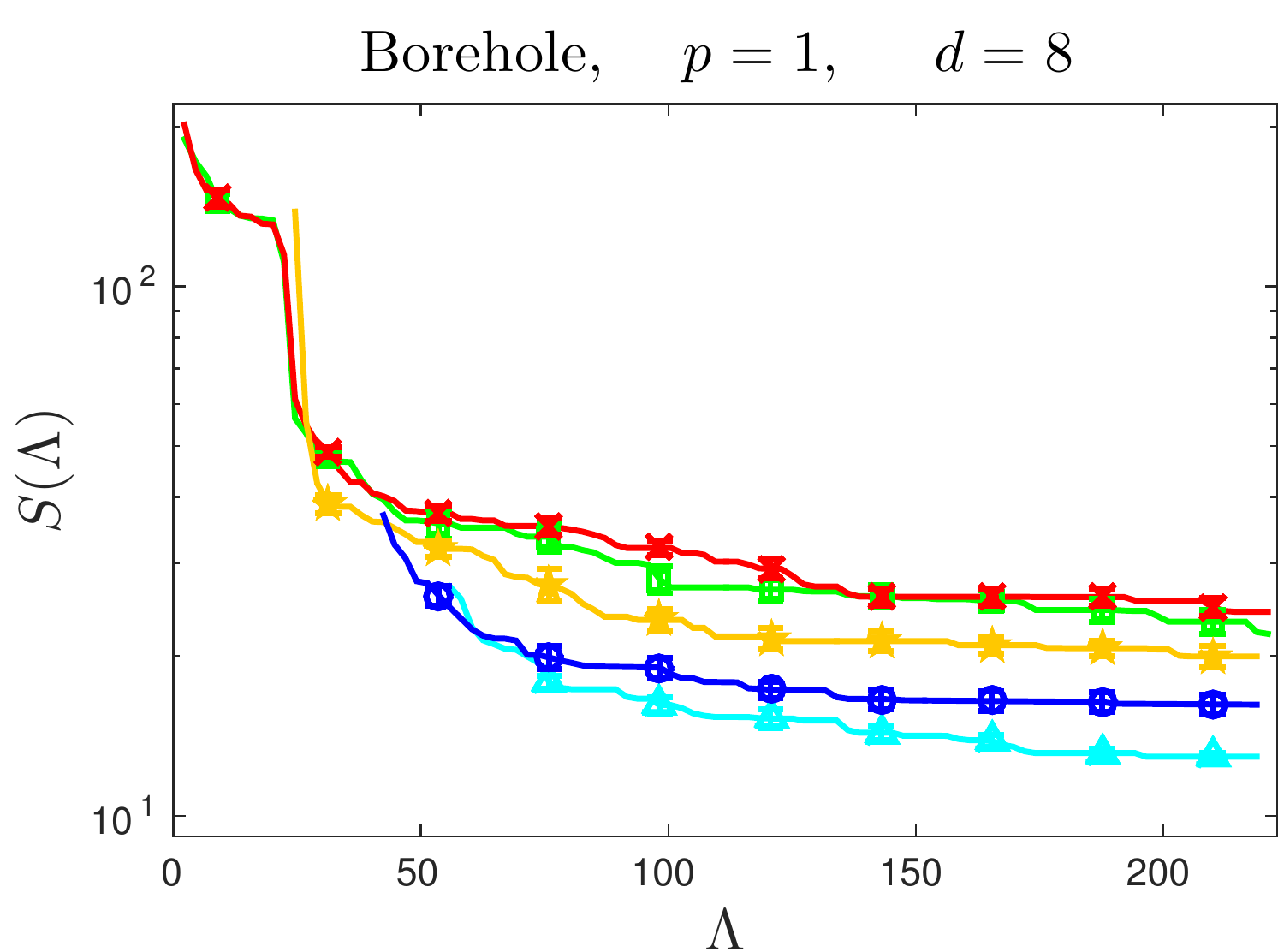} \hspace{\imhspthree}
% } \\[\imcaptionspace]
} \\[-0.15in]
\caption[]{\small
% The simple regret $S(\capital)$ against capital $\capital$ on $6$ synthetic problems.
Results on $6$ synthetic problems where we plot the simple regret $S(\capital)$ (lower is
better) against the capital $\capital$.
The title states the function used, and the fidelity and domain dimesions.
For the first two figures we used capital $30\cost(\zhf)$, therefore a method which
only queries at $g(\zhf,\cdot)$ can make at most $30$ evaluations.
For the third figure we used $50\cost(\zhf)$, for the fourth $100\cost(\zhf)$ and for
the last $200\cost(\zhf)$ to reflect the dimensionality $d$ of $\Xcal$.
The curves for the multi-fidelity methods start mid-way since they have not queried
at $\zhf$ up until that point.
All curves were produced by averaging over $20$ experiments and the error bars
indicate one standard error.
\vspace{\imtextspace}
\label{fig:toy}
}
\end{figure*}
}

\newcommand{\insertRealResultsFigure}{
\begin{figure*}
\centering
\hspace{\imleftspace}
\subfigure[]{
  \includegraphics[width=\imarrwthree]{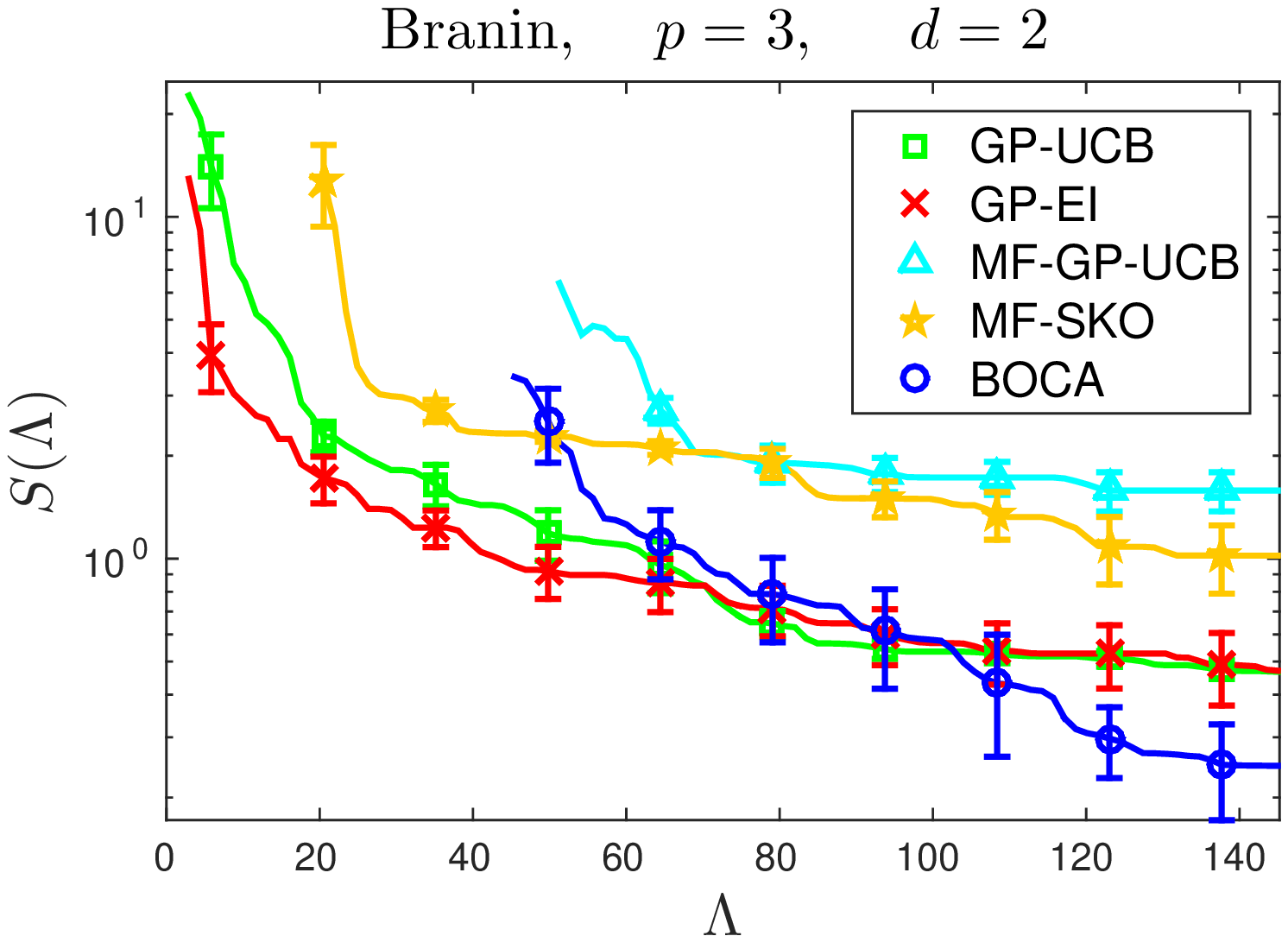} \hspace{\imhspthree}
  \label{fig:branin}
}
\subfigure[]{
  \includegraphics[width=\imarrwthree]{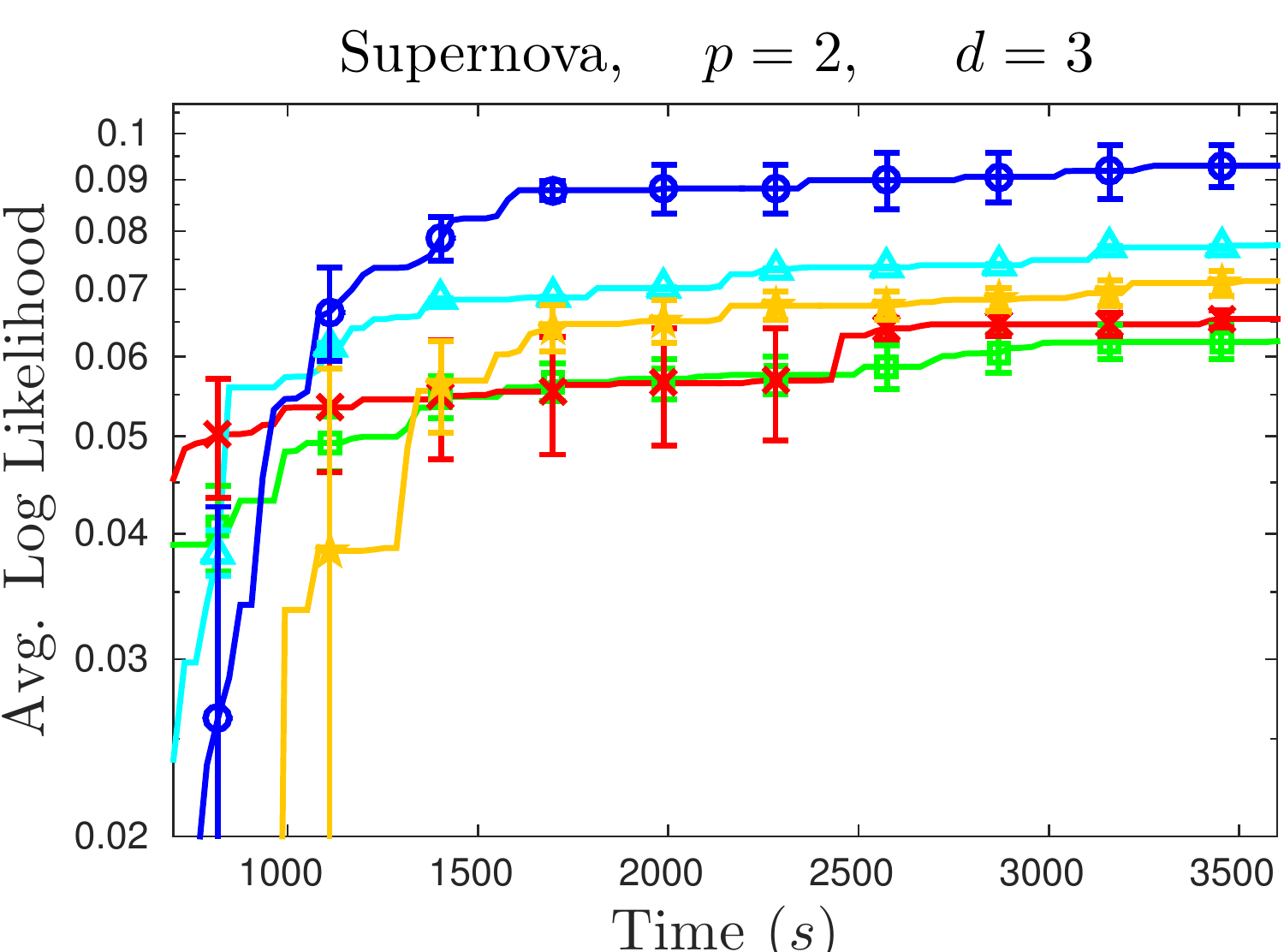} \hspace{\imhspthree}
  \label{fig:sn}
}
\subfigure[]{
  \includegraphics[width=\imarrwthree]{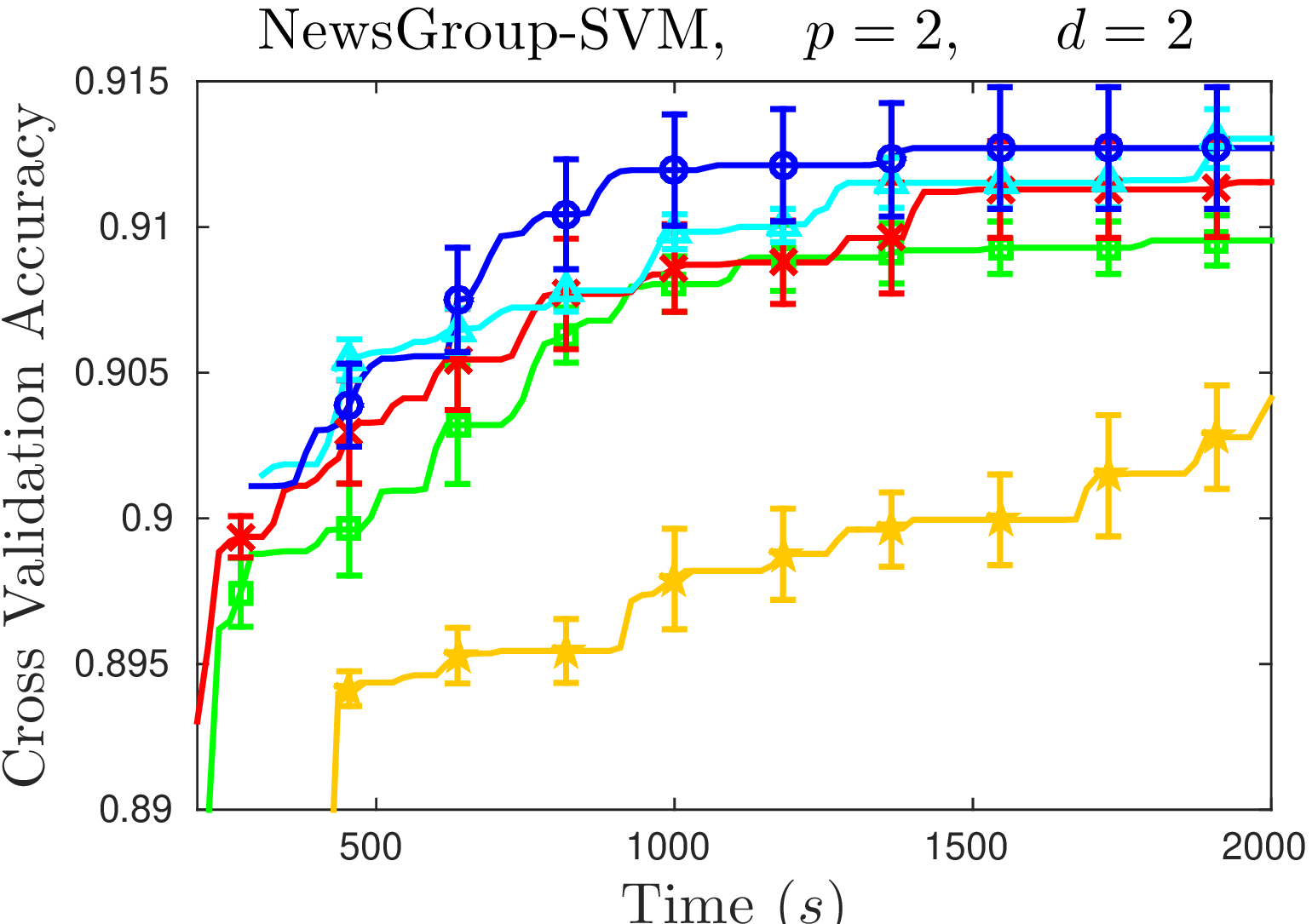} \hspace{\imhspthree}
  \label{fig:news}
% } \\[\imcaptionspace]
} \\[-0.15in]
\caption[]{\small
\subref{fig:branin}: The synthetic benchmark with the Branin function where we used a
capital of $50\cost(\zhf)$. See caption under Fig.~\ref{fig:toy} for more details.
\subref{fig:sn},~\subref{fig:news}: Results on the supernova and news group experiments
from sections~\ref{sec:sn} and~\ref{sec:news} respectively. We have plotted the maximum
value (higher is better) against wall clock time.
~\subref{fig:branin} was averaged over $20$ experiments while~\subref{fig:sn}
and~\subref{fig:news} were averaged over $10$ xperiments each.
The error bars indicate one standard error.
\vspace{\imtextspace}
\label{fig:real}
}
\end{figure*}
}

\newcommand{\insertAlgoMFGPUCBTwo}{
% \vspace{-0.05in}
\begin{algorithm}
\textbf{Input: } kernel $\kernel$.
\vspace{-0.07in}
\begin{itemize}[leftmargin=0.18in]
\item Set $\;\;\nutt{0}(\cdot)\leftarrow\zero$,
          $\;\;\tautt{0}(\cdot)\leftarrow\kernel(\cdot,\cdot)^{1/2}$, 
          $\;\;\Dcal_0\leftarrow\emptyset$. 
\vspace{-0.05in}
\item for $t = 1, 2,\dots$
\vspace{-0.05in}
  \begin{enumerate}[leftmargin=0.18in]
  \item $\xt \leftarrow \argmax_{x\in\Xcal} \,\utilt(x)$. 
  \hfill See~\eqref{eqn:mfucb}
  \item $\zt \leftarrow \argmin_{z\in\,\candfidelst(\xt)\cup\{\zhf\}} \,\cost(z)$. 
  \hfill See~\eqref{eqn:candfidelst}
  \item $\yt \leftarrow$ Query $\gunc$ at $(\zt,\xt)$.
  \item $\Dcal_t\leftarrow\Dcal_{t-1}\cup\{(\zt,\xt,\yt)\}$.
        Update posterior mean $\nutt{t}$, and standard deviation $\tautt{t}$ 
        for $\gunc$ conditioned on $\Dcal_t$.
%   \item \textbf{for} $m=1,\dots,M$:  \\
%     \vphantom{T} \hspace{0.1in} 
%       \textbf{if} $\betath\sigmamtmo(\xt) > \gamma$, \textbf{break}; \\
%     $\mt = m$.
%   \item $\Dcalmtt \leftarrow \Dcalmmtt{\mt}{t-1}\cup\{(\xt,\yt)\}$.
%   \item $\Dcalmt \leftarrow \Dcalmmtt{m}{t-1}$ for all $m\neq \mt$.
%   \item Obtain $\mumtt,\sigmamtt$ conditioned on $\Dcalmtt$.
  \end{enumerate}
% \vspace{-0.1in}
\end{itemize}
\vspace{-0.1in}
\caption{\hspace{0.05in}\mfgpucbtwo \label{alg:mfgpucbtwo}}
\end{algorithm}
% \vspace{-0.05in}
}

% Algorithm gpbbalgo in Appendix 
%%%%%%%%%%%%%%%%%%%%%%%%%%%%%%%%%%%%%%%%%%%%%%%%%%%%%%%%%%%%%%%%%%%%%%%%%%%%%%%%
\newcommand{\insertAlgorithmGPBALGO}{
\begin{algorithm}
\vspace{0.02in}
\textbf{Input: }
kernel $\kernel$.
\vspace{-0.10in}
\begin{itemize}%[label={}]
\item $\Dcal_0 \leftarrow \emptyset$,
  $(\mu_0, \sigma_0) \leftarrow (\zero, \kernel^{1/2})$.
\vspace{-0.05in}
\item \textbf{for} $t=1, 2, \dots$
\vspace{-0.13in}
\begin{enumerate}
  \item $\xt \leftarrow \argmax_{x\in \Xcal} \mu_{t-1}(x) +
          \betath\sigma_{t-1}(x)$
  \item $\yt \leftarrow \textrm{Query $\func$ at $\xt$}$.
%   \item $\Dcal_{t} = \Dcal_{t-1} \cup \{ (\xt, \yt)\}$.
  \item Perform Bayesian posterior updates to obtain $\mu_{t},
    \sigma_{t}$ \hfill See~\eqref{eqn:gpPost}.
\end{enumerate}
\vspace{-0.15in}
\end{itemize}
\caption{$\;$\gpucb \label{alg:gpucb}\hfill\citep{srinivas10gpbandits}$\quad$}
\end{algorithm}
}

\begin{abstract}
Bandit methods for black-box optimisation, such as Bayesian optimisation, 
are used in a variety of applications including hyper-parameter tuning and
experiment design.
Recently, \emph{multi-fidelity} methods have garnered
considerable attention since function evaluations have become increasingly expensive in
such applications.
Multi-fidelity methods use cheap approximations to the function of
interest to speed up the overall optimisation process.
However, most multi-fidelity methods assume only a finite number of approximations.
In many practical applications however, a continuous spectrum of approximations might be
available.
For instance, when tuning an expensive neural network, one might choose to approximate the
cross validation performance using less data $N$ and/or few training iterations $T$.
Here, the approximations are best viewed as arising out of a continuous two dimensional
space $(N,T)$.
In this work, we develop a Bayesian optimisation method, \boca, for this setting.
We characterise its theoretical properties and show that it achieves better regret than
than strategies which ignore the approximations.
\bocas outperforms several other baselines in synthetic and real experiments.
\end{abstract}

\section{Introduction}
\label{sec:intro}

Many tasks in scientific and engineering applications can be framed as \emph{bandit
optimisation} problems, where we need to sequentially evaluate a noisy
black-box function $\func:\Xcal\rightarrow\RR$ with the goal of finding its optimum.
Some applications include hyper-parameter tuning in machine learning~\citep{%
snoek12practicalBO,hutter11smac}, optimal policy search~\citep{lizotte07automaticGait,
martinez07robotplanning} and scientific experiments~\citep{parkinson06wmap3,gonzalez14gene}.
Typically, in such applications, each function evaluation is expensive and historically,
the bandit literature has focused on developing methods for finding the optimum
while keeping the number of evaluations to $\func$ at a minimum.

However, with increasingly expensive function evaluations, conventional methods have become
infeasible as a significant cost needs to be expended before we can learn anything
about $\func$.
As a result, \emph{multi-fidelity} optimisation methods have recently gained
attention~\citep{kandasamy16mfbo,li2016hyperband,cutler14mfsim}.
As the name suggests, these methods assume that we have access to lower fidelity
approximations to $\func$ which can be evaluated instead of $\func$.
The lower the fidelity, the cheaper the evaluation, but it provides less accurate
information about $\func$.
For example, when optimising the configuration of an expensive real world robot,
its performance can be approximated using cheaper computer simulations.
The goal is to use the cheap approximations to guide search for the optimum of $\func$,
and reduce the overall cost of optimisation.
However, most multi-fidelity work assume only a finite number of approximations.
In this paper, we study multi-fidelity optimisation when there is access to a continuous
spectrum of approximations.

To motivate this,
consider tuning a classification algorithm over a space of hyper-parameters $\Xcal$
by maximising a validation set accuracy. 
The algorithm is to be trained using $\Nmax$
data points via an iterative algorithm for $\Tmax$ iterations.
However, we wish to use fewer training points $N<\Nmax$ and/or fewer iterations
$T<\Tmax$ to approximate the validation accuracy.
We can view validation accuracy as a function
$g:[1,\Nmax]\times[1,\Tmax]\times\Xcal\rightarrow\RR$ where evaluating $g(N,T,x)$
requires training the algorithm with $N$ points for $T$ iterations with the
hyper-parameters $x$.
If the training complexity of the algorithm is quadratic in data size and linear in the
number of iterations, then the cost of this evaluation is $\lambda(N,T) = \bigO(N^2T)$.
Our goal is to find the optimum when $N=\Nmax$, and $T=\Tmax$,
% i.e.  $\xopt = \argmax_{x\in\Xcal} g(\Nmax,\Tmax, x)$.
i.e.  we wish to maximise $\func(x) = g(\Nmax,\Tmax, x)$.

% In this setting, the approximations may be viewed as coming from a continuous $2$
In this setting, while $N,T$ are technically discrete choices, they are more naturally
viewed as coming from a continuous $2$
% In this setting, the approximations are more naturally viewed 
dimensional \emph{fidelity space}, $[1, \Nmax]\times[1,\Tmax]$.
% Since querying $g$ at $N,T$ values less than $\Nmax,\Tmax$ costs less, we are
% tempted to use such queries in the hope that we may
One might hope that cheaper queries to $g(N,T,\cdot)$ with $N,T$ less than
$\Nmax,\Tmax$ can be used to learn about $g(\Nmax, \Tmax,\cdot)$ and consequently
optimise it using less overall cost.
Indeed, this is the case with many machine learning algorithms where cross
validation performance tends to vary smoothly with data set size and number of iterations.
Therefore, one may use cheap low fidelity experiments with small $(N, T)$ to discard
bad hyper-parameters and deploy expensive high fidelity experiments with large
$(N,T)$ only in a small but promising region.
The main theoretical result of this paper (Theorem~\ref{thm:regret}) shows that our proposed algorithm, \boca{}, exhibits precisely this behaviour. 

Continuous approximations also arise in
simulation studies:
% where simulations can be carried out at varying levels of granularity,
where simulations can be carried out at varying levels of granularity,
on-line advertising: where an ad can be controlled by continuous
parameters such as display time or target audience, and several other
experiment design tasks.
In fact, in many multi-fidelity papers, the finite approximations were
obtained by discretising
a continuous space~\citep{kandasamy16mfbo,huang06mfKriging}.
Here, we study a Bayesian Optimisation technique that is directly designed for
continuous fidelity spaces and
% could potentially be applied to more general spaces.
is potentially applicable to more general spaces.
Our \textbf{main contributions} are,\hspace{-0.1in}
\vspace{-0.15in}
\begin{enumerate}
\item A novel setting and model for multi-fidelity optimisation with continuous
approximations using Gaussian process (GP) assumptions.
We develop a novel algorithm, \boca, for this setting.
\vspace{-0.10in}
\item A theoretical analysis characterising the behaviour and regret bound for \boca.
\vspace{-0.10in}
\item An empirical study which demonstrates that \bocas outperforms alternatives,
  both multi-fidelity and otherwise,
  on a series of synthetic problems and real examples in hyper-parameter tuning
  and astrophysics.
%   Our python implementation will be made publicly 
\end{enumerate}

\vspace{-0.17in}

\subsection*{Related Work}
\vspace{-0.07in}

Bayesian optimisation (\bayo), refers to a suite of techniques for bandit optimisation
which use a prior belief distribution for $\func$.
While there are several techniques for
BO~\citep{mockus94bo,jones98expensive,thompson33sampling,lobato14pes,defreitas12expregret},
our work will build on the 
Gaussian process upper confidence bound (\gpucb) algorithm
of~\citet{srinivas10gpbandits}.
\gpucbs models $\func$ as a GP and uses upper confidence bound (UCB)~\citep{auer03ucb}
techniques to determine the next point for evaluation.

\bayo{} techniques have been used in developing multi-fidelity optimisation methods
in various applications such as hyper-parameter tuning and industrial
design~\citep{huang06mfKriging,swersky2013multi,klein2015towards,forrester07cokriging,%
poloczek2016multi}.
However, these methods are either problem specific and/or only use a finite
number of fidelities.
Further, none of them come with theoretical underpinnings.
Recent work has studied multi-fidelity methods for specific problems 
such as hyper-parameter tuning, active learning and
reinforcement learning%
~\citep{agarwal2011oracle,sabharwal2015selecting,cutler14mfsim,zhang15weakAndStrong,%
li2016hyperband}.
While some of the above tasks can be framed as optimisation problems, the methods
themselves are specific to the problem considered.
Our method is more general as it applies to any bandit optimisation task.

Perhaps the closest work to us is that of~\citet{kandasamy16mfbo,kandasamy16mfbandit}
who developed \mfgpucbones assuming a finite number of approximations to $\func$.
% This line of work was important as it provided the first theoretical guarantees for
% multi-fidelity optimisation.
While this line of work was the first to provide theoretical guarantees for
multi-fidelity optimisation, it has two important shortcomings.
First, they make strong assumptions, particularly a uniform bound on the
difference between the expensive function and an approximation.
This does not allow for instances where an approximation might be good at certain regions
but not at the other.
In contrast, our probabilistic treatment between fidelities is is robust
to such cases.
Second, their model does not allow sharing information between fidelities; each
approximation is treated independently.
% In addition to not uti
Not only is this wasteful as lower fidelities can provide useful information about
higher fidelities, it also means that the algorithm might perform poorly if the fidelities
are not designed properly.
We demonstrate this with an experiment in Section~\ref{sec:experiments}.
On the other hand, our model allows sharing information across the fidelity space in a
natural way.
In addition, we can also handle continuous approximations whereas their method is strictly
for a finite number of approximations.
That said, \mfgpucbtwos inherits a key intuition from \mfgpucbone,
% \todo{we don't need ``-I'' anymore right? Not reading the intro part too closely yet}
% first select the next point for evaluation using an upper confidence bound strategy;
which is to choose a fidelity only if we have sufficiently reduced the uncertainty at
all lower fidelities.
% However, in addition to dealing with continuous approximations, we also differ from%
% ~\citet{kandasamy16mfbo} fundamentally in our model assumptions.
% Besides this, the mechanics
% of our algorithm and most analysis techniques are considerably different.
Besides this, there are considerable differences in the mechanics
of the algorithm and proof techniques.
As we proceed, we will draw further comparisons to~\citet{kandasamy16mfbo}.

\section{Preliminaries}
\label{sec:prelims}

% \subsection{A Review of Gaussian processes and \gpucb}
\subsection{Some Background Material}
\label{sec:prelims}

\textbf{Gaussian processes:}
A GP over a space $\Xcal$ is a random process from $\Xcal$ to $\RR$.
GPs are typically
used as a prior for functions in Bayesian nonparametrics.
It is characterised by a mean function $\mu:\Xcal\rightarrow\RR$ and
a covariance function (or kernel) $\kernel:\Xcal^2\rightarrow\RR$.
If $\func\sim\GP(\mu,\kernel)$, then $f(x)$ is distributed normally
$\Ncal(\mu(x), \kernel(x,x))$ for all $x\in\Xcal$.
Suppose that we are given $n$ observations $\Dcal_n=\{(x_i, y_i)\}_{i=1}^n$ from this GP,  where
$x_i\in\Xcal$, $y_i = \func(x_i) + \epsilon_i \in\RR$ and $\epsilon_i\sim\Ncal(0,\eta^2)$.
Then the posterior process $\func|\Dcal_n$ is also a GP with mean $\mu_n$ and
covariance $\kernel_n$ given by 
\begin{align*}
\hspace{-0.05in}
% \mu_n(x) &= \kb^\top\Delta^{-1}Y, \hspace{0.35in}
% \numberthis \label{eqn:gpPost} \\
% \kernel_n(x,x') &= \kernel(x,x') - \kb^\top\Delta^{-1}\kb',
\mu_n(x) &= k^\top(K + \eta^2I)^{-1}Y, \hspace{0.35in}
\numberthis \label{eqn:gpPost} \\
\kernel_n(x,x') &= \kernel(x,x') - k^\top(K + \eta^2I)^{-1}k',
\hspace{0.05in}
\end{align*}
where $Y\in\RR^n$ is a vector with $Y_i=y_i$, and
$k,k'\in\RR^n$ are such that $k_i =
\kernel(x,x_i),k'_i=\kernel(x',x_i)$.
% Also, the matrix $\Delta\in \RR^{n\times n}$ is defined as 
% $\Delta = \Kb + \eta^2I$ where
% $\Kb_{i,j} = \kernel(x_i,x_j)$.
The matrix $K\in \RR^{n\times n}$ is given by
$K_{i,j} = \kernel(x_i,x_j)$.
We refer the reader to chapter 2 of~\citet{rasmussen06gps} for more on the basics of
GPs and their use in regression.
% 
% \begin{align*}
% \hspace{-0.05in}
% % \mu_n(x) &= \kb^\top\Delta^{-1}Y, \hspace{0.35in}
% % \numberthis \label{eqn:gpPost} \\
% % \kernel_n(x,x') &= \kernel(x,x') - \kb^\top\Delta^{-1}\kb',
% \mu_n(x) &= \kb^\top(\Kb + \eta^2I)^{-1}Y, \hspace{0.35in}
% \numberthis \label{eqn:gpPost} \\
% \kernel_n(x,x') &= \kernel(x,x') - \kb^\top(\Kb + \eta^2I)^{-1}\kb',
% \hspace{0.05in}
% \end{align*}
% where $Y\in\RR^n$ is a vector with $Y_i=y_i$, and
% $\kb,\kb'\in\RR^n$ are such that $\kb_i =
% \kernel(x,x_i),\kb'_i=\kernel(x',x_i)$.
% % Also, the matrix $\Delta\in \RR^{n\times n}$ is defined as 
% % $\Delta = \Kb + \eta^2I$ where
% % $\Kb_{i,j} = \kernel(x_i,x_j)$.
% The matrix $\Kb\in \RR^{n\times n}$ is given by
% $\Kb_{i,j} = \kernel(x_i,x_j)$.
% We refer the reader to chapter 2 of~\citet{rasmussen06gps} for more on the basics of
% GPs and their use in regression.
% % We refer the reader to~\citet{rasmussen06gps} for more on GPs.

\textbf{Radial kernels:}
The prior covariance functions of GPs are typically taken to be \emph{radial kernels};
% The squared exponential (SE) kernel and \matern{} kernels, which are popularly used in
% practice, are radial kernels.
some examples are the squared exponential (SE)
and \matern{} kernels. 
Using a radial kernel means that the prior covariance can be written as
$\kernel(x,x') = \kernelscale\phi(\|x-x'\|)$ and
depends only on the distance between $x$ and $x'$.
Here, the scale parameter $\kernelscale$ 
captures the magnitude $\func$ could deviate from $\mu$.
The function $\phi:\RR_{+}\rightarrow\RR_+$ is a decreasing function with
$\|\phi\|_\infty = \phi(0) = 1$.
In this paper, we will use the SE kernel in a running example to convey the
intuitions in our methods. 
For the SE kernel, $\phi(r) = \radialh(r) = \exp(-r^2/(2h^2))$, where $h\in\RR_+$,  called the \emph{bandwidth} of the kernel, controls the smoothness of the GP.
% It is well known that 
When $h$ is large, the samples drawn from the GP tend to
be smoother as illustrated in Fig.~\ref{fig:gpBWSample}.
We will reference this observation frequently in the text.

\insertFigBWSample

\textbf{\gpucb:}
The Gaussian Process Upper Confidence Bound (\gpucb) algorithm
of~\citet{srinivas10gpbandits}  is a method for bandit optimisation,
which, like many other \bayos{} methods, models $\func$ as a sample from a Gaussian process.
At time $t$, the next point $\xt$ for evaluating $\func$ is chosen via the following
procedure.
First, we construct an upper confidence bound
$\utilt(x) = \mutmo(x) + \betath \sigmatmo(x)$ for the GP. 
$\mutmo$ is the posterior mean of the GP conditioned on the previous $t-1$ evaluations
and $\sigmatmo$ is the posterior standard deviation.
Following other UCB algorithms~\citep{auer03ucb},
the next point is chosen by maximising $\utilt$, i.e.
$\xt=\argmax_{x\in\Xcal}\utilt(x)$.
The $\mutmo$ term encourages an \emph{exploitative} strategy -- in that we want to query
regions where we already believe  $\func$ is high -- and $\sigmatmo$ encourages
an \emph{exploratory} strategy -- in that we want to query where we are uncertain
about $\func$ so that we do not miss regions which have not been queried yet.
$\betat$, which is typically increasing with $t$, controls the trade-off between
exploration and exploitation.
We have provided a brief review of \gpucbs in Appendix~\ref{sec:gpucbReview}.

% \vspace{-0.05in}

\subsection{Problem Set Up}
% \vspace{-0.05in}
\label{sec:prelims}

% In stochastic bandit optimisation, our goal is to maximise a function
Our goal in bandit optimisation is to maximise a function
$\func:\Xcal\rightarrow \RR$, over a domain $\Xcal$.
When we evaluate $\func$ at $x\in\Xcal$ we observe $y=\func(x) + \epsilon$ where
$\EE[\epsilon] = 0$.
Let $\xopt \in \argmax_{x\in\Xcal}\func(x)$ be a maximiser of $\func$ and
$\funcopt = \func(\xopt)$ be the maximum value.
% $\func$ is accessible only via noisy point evaluations and has no gradient information.
An algorithm for bandit optimisation is a sequence of points $\{\xt\}_{t\geq0}$, where
at time $t$, the algorithm chooses to evaluate $\func$ at $\xt$ based on previous queries
and observations $\{(\xtt{i},\ytt{i})\}_{i=1}^{t-1}$.
After $n$ queries to $\func$,
its goal is to achieve small \emph{simple regret} $\Sn$, as defined below.
\begin{align*}
\Sn = \min_{t=1,\dots,n} \funcopt - \func(\xt).
\numberthis
\label{eqn:SnDefn}
\end{align*}
% \textbf{Continuous approximations in multi-fidelity optimisation:}
% \textbf{Multi-fidelity Bandit Optimisation:}
\textbf{Continuous Approximations:}
In this work, we will let $\func$ be a slice of a function $\gunc$ that lies
in a larger space.
Precisely, we will assume the existence of a fidelity space $\Zcal$ and a function 
$\gunc:\Zcal\times\Xcal \rightarrow \RR$ defined on the product space of the fidelity
space and domain.
The function $\func$ which we wish to maximise is related to $\gunc$ via
$\func(\cdot) = \gunc(\zhf, \cdot)$, where $\zhf\in\Zcal$.
For instance, in the hyper-parameter tuning example from Section~\ref{sec:intro},
$\Zcal = [1,\Nmax]\times[1,\Tmax]$ and $\zhf = [\Nmax, \Tmax]$.
Our goal is to find a maximiser $\xopt \in \argmax_x \func(x) = \argmax_x \gunc(\zhf, x)$.
We have illustrated this setup in Fig.~\ref{fig:fidelSpace}.
In the rest of the manuscript, the term ``fidelities'' will refer to points $z$
in the fidelity space $\Zcal$.
\insertFigFidelSpace

The multi-fidelity framework is attractive when the following two conditions are true
about the problem.
\vspace{-0.10in}
% \begin{itemize}[label=-\hspace{-0.02in},leftmargin=0.13in]
\begin{enumerate}[leftmargin=0.19in]
\item
\ul{There exist fidelities $z\in\Zcal$ where evaluating $\gunc$ is cheaper than
evaluating at $\zhf$.}
To this end, we will associate a \emph{known} cost function $\cost:\Zcal\rightarrow
\RR_+$.
In the hyper-parameter tuning example, $\cost(z) = \cost(N,T) = \bigO(N^2T)$.
It is helpful to think of $\zhf$ as being the most expensive fidelity,
i.e. maximiser of $\cost$, and that $\cost(z)$ decreases as we move away from
$\zhf$. 
% However, our algorithm
% and results apply to much more general situations.\hspace{-0.1in}
However, this notion is strictly not necessary for our algorithm or results.\hspace{-0.1in}
\vspace{-0.06in}
\item
\ul{The cheap $\gunc(z,\cdot)$ evaluation gives us information about
$g(\zhf,\cdot)$.}
This is true if $\gunc$ is smooth across the fidelity space
as illustrated in Fig.~\ref{fig:fidelSpace}.
As we will describe shortly, this smoothness can be achieved by modelling $\gunc$ as
a GP
with an appropriate kernel for the fidelity space $\Zcal$.
% \end{itemize}
\end{enumerate}
\vspace{-0.12in}

In the above setup, a multi-fidelity algorithm is a sequence of query-fidelity pairs
$\{(\zt,\xt)\}_{t\geq 0}$
where, at time $t$, the algorithm chooses $\zt\in\Zcal$ and $\xt\in\Xcal$, and observes
$\ytt{t} = g(\ztt{t},\xtt{t}) + \epsilon$ where $\EE[\epsilon] = 0$.
The choice of $(\zt, \xt)$ can of course depend on the 
previous fidelity-query-observation triples $\{(\ztt{i},\xtt{i},\ytt{i})\}_{i=1}^{t-1}$.

\textbf{Multi-fidelity Simple Regret:}
We provide bounds on the simple regret $S(\capital)$ of a multi-fidelity optimisation method
after it has spent capital $\capital$ of a resource.
Following~\citet{srinivas10gpbandits}, we
will aim to provide \emph{any capital} bounds, meaning that an algorithm would be expected
to do well for \emph{all} values of (sufficiently large) $\capital$.
Say we have made $N$ queries to $g$ within capital $\capital$,
i.e. $N$ is the \emph{random} quantity such that
$N = \max\{n\geq 1: \sum_{t=1}^n \cost(\zt) \leq \capital\}$.
While the cheap evaluations at $z\neq \zhf$ are useful in guiding search 
for the optimum of $\gunc(\zhf,\cdot)$, there is no reward for optimising a cheaper
$g(z,\cdot)$.
Accordingly, we define the simple regret after capital $\capital$ as,
\vspace{-0.02in}
\begin{align*}
S(\capital) =
\begin{cases}
% \displaystyle\min_{\substack{t=\,1,\dots,N\\t\,:\,\zt = \zhf}}\;
\displaystyle\min_{\substack{t\in\{1,\dots,N\}\\\text{s.t}\;\zt = \zhf}}
  \funcopt - \func(\xt) % \hspace{0.01in}
    &\text{if we have queried at $\zhf$,} \\
+\,\infty & \text{otherwise.}
\end{cases}
\\[-0.15in]
% \numberthis
% \label{eqn:SLambdaDefn}
\end{align*}
\vspace{-0.35in}

This definition reduces to the single fidelity definition~\eqref{eqn:SnDefn} when we only
query $g$ at $\zhf$.
It is also similar to the definition in~\citet{kandasamy16mfbo}, but unlike them, we do
not impose additional boundedness constraints on $\func$ or $\gunc$.

Before we proceed, we note that it is customary in the bandit literature to analyse
\emph{cumulative regret}. However, the definition of cumulative regret depends
on the
application at hand~\citep{kandasamy16mfbandit} and the results in this paper can be
extended to to many sensible notions of cumulative regret.
However, both to simplify exposition and since our focus in this paper is optimisation,
we stick to simple regret.

\textbf{Assumptions:}
As we will be primarily focusing on continuous and compact domains and fidelity spaces,
going forward we will assume, without any loss of generality, that $\Xcal = [0, 1]^d$ and
$\Zcal = [0, 1]^p$.
We discuss non-continuous settings briefly at the end of Section~\ref{sec:mfgpucb}.
In keeping with similar work in the Bayesian optimisation literature, we will
assume
$\gunc\sim\GP(\zero,\kernel)$ and upon querying at $(z,x)$ we observe
$y = g(z,x) + \epsilon$ where $\epsilon\sim\Ncal(0,\eta^2)$.
$\kernel:(\Zcal\times\Xcal)^2\rightarrow \RR$ is
the prior covariance defined on the product space.
In this work, we will study exclusively $\kernel$ of the following form,
\begin{align*}
\hspace{-0.1in}
\kernel([z,x],[z',x']) \,=\,
\kernelscale\,\phiz(\|z-z'\|)\, \phix(\|x-x'\|).
\label{eqn:mfkernel}
\numberthis
\end{align*}
Here, $\kernelscale\in\RR_+$ is the scale parameter and $\phiz,\phix$ are radial kernels
defined on $\Zcal,\Xcal$ respectively.
The fidelity space kernel $\phiz$ is an important component in this work.
It controls the
smoothness of $\gunc$ across the fidelity space and hence determines how much
information the lower fidelities provide about $g(\zhf,\cdot)$.
For example, suppose that $\phiz$ was a SE kernel.
A favourable setting for
a multi-fidelity method would be for $\phiz$ to have a large bandwidth $\hz$
as that would imply that $g$ is very smooth across $\Zcal$.
We will see that $\hz$ determines the behaviour and theoretical guarantees
of \mfgpucbtwos in a natural way when $\phiz$ is the SE kernel.
% To formalise this, we will  define the following function
To formalise this notion, we will  define the following function
$\inffunc:\Zcal\rightarrow[0,1]$.
\begin{align*}
\inffunc(z) = \sqrt{1 - \phiz(\|z-\zhf\|)^2}
\label{eqn:inffunc}
\numberthis
\end{align*}
One interpretation of $\inffunc(z)$ is that it measures the \emph{gap} in
information about $g(\zhf,\cdot)$ when we query at $z\neq\zhf$.
That is, it is the price we have to pay, in information, for querying at a cheap fidelity.
Observe that $\inffunc$ increases when we move away from $\zhf$ in the fidelity space.
For the SE kernel, it can be shown\footnote{%
Strictly, $\inffunc(z) \leq \|z-\zhf\|/\hz$, but the inequality is  tighter for
larger $\hz$.
In any case, $\inffunc$ is strictly decreasing with $\hz$.
}
$\inffunc(z) \approx \frac{\|z-\zhf\|}{\hz}$.
For large $\hz$, $\gunc$ is smoother across $\Zcal$ and we can expect the lower fidelities
to be more informative about $\func$;  as expected the information gap $\inffunc$ is
small for large $\hz$.
If $\hz$ is small and $\gunc$ is not smooth, the gap $\inffunc$ is large
and lower fidelities are not as informative.

% \todo{@samy, see comment in the code below}
%%% @samy: when defining functions, dont forcibly include space. The right way to do it is to define the command as usual and call the function with trailing parentheses. This is because Latex, by default, assumes that any character following \bayos is an argument to the function \bayos, by default. This is why it messes up spaces sometimes. So, the right way to do it is to always call functions as '\bayos{}' (I have done this in this case).}

Before we present our algorithm for the above setup, we will introduce notation for the posterior GPs for $\gunc$ and $\func$.
Let $\Dcal_n = \{(\ztt{i},\xtt{i},\ytt{i})\}_{t=1}^n$ be $n$ fidelity, query,
observation values from the GP $\gunc$, where $\ytt{i}$ was observed when evaluating
$g(\ztt{i},\xtt{i})$.
We will denote the posterior mean and standard
deviation of $\gunc$ conditioned on $\Dcal_n$ by $\nutt{n}$ and $\tautt{n}$ respectively
($\nutt{n},\tautt{n}$ can be computed from~\eqref{eqn:gpPost} by replacing
$x\leftarrow[z,x]$).
Therefore $\gunc(z,x)|\Dcal_n\sim\Ncal(\nutt{n}(z,x), \tausqtt{n}(z,x))$ for all
$(z,x)\in\Zcal\times\Xcal$.
We will further denote 
\begin{align*}
\mutt{n}(\cdot) = \nutt{n}(\zhf,\cdot), \hspace{0.2in}
\sigmatt{n}(\cdot) = \tautt{n}(\zhf,\cdot),
\label{eqn:munsigman}
\numberthis
\end{align*}
to be the posterior mean and standard
deviation of $\gunc(\zhf,\cdot) = \func(\cdot)$.
It follows that $\func|\Dcal_n$ is also a GP and satisfies
$\func(x)|\Dcal_n \sim \Ncal(\mutt{n}(x), \sigmasqtt{n}(x))$ for all $x\in\Xcal$.

\section{\mfgpucbtwo: Bayesian Optimisation with Continuous Approximations}
\label{sec:mfgpucb}

% We now present our \mfgpucbtwos algorithm.
\mfgpucbtwos is a sequential strategy to select a domain point $\xt\in\Xcal$ and
fidelity $\zt\in\Zcal$ at time $t$ based on previous observations.
At time $t$, we will first
construct an upper confidence bound $\utilt$ for the function $\func$ we wish to optimise.
It takes the form,
\begin{align*}
\utilt(x) = \mutmo(x) + \betath \sigmatmo(x).
\numberthis
\label{eqn:mfucb}
\end{align*}
Recall from~\eqref{eqn:munsigman} that $\mutmo$ and $\sigmatmo$ are the posterior mean and
standard deviation of $\func$
% i.e. $\gunc$ restricted to $\zhf$,
using the observations from the previous $t-1$ time steps
at all fidelities, i.e. the entire $\Zcal\times\Xcal$ space.
% in the entire $\Zcal\times\Xcal$ space.
We will specify $\betat$ in theorems~\ref{thm:regret},~\ref{thm:main}. 
Following other UCB algorithms, our next point $\xt$ in the domain $\Xcal$ for evaluating
$\gunc$ is a maximiser of $\utilt$, i.e. $\xt \in \argmax_{x\in\Xcal} \utilt(x)$.

Next, we need to determine the fidelity $\zt\in\Zcal$ to query $g$.
For this we will first select a subset $\candfidelst(\xt)$ of $\Zcal$ as follows,
\begingroup
% \allowdisplaybreaks
\begin{align*}
&\candfidelst(\xt) = 
%   \{\zhf\}\cup\big
    \Big\{z\in\Zcal\,:\,\;\cost(z) < \cost(\zhf),
    \quad\tautmo(z,\xt) > \gamma(z), \\
%     &\hspace{1.25in}\inffunc(z) > \frac{\inffunc(\diamz) }{\betath}
%     &\hspace{1.25in}\inffunc(z) > \inffunc(\diamz) /\betath
    &\hspace{1.25in}\inffunc(z) > \betatmh\inffunc(\diamz)
      \Big\}, 
\numberthis\label{eqn:candfidelst}
\\[0.05in]
&\textrm{where } \quad\gamma(z) =\,
      \sqrt{\kernelscale}\;
        \inffunc(z)\;
        \bigg(\hspace{-0.02in}\frac{\cost(z)}{\cost(\zhf)}
%         \hspace{-0.02in}\bigg)^{\frac{1}{d+p+2}}
        \hspace{-0.02in}\bigg)^{q}
\,.
\end{align*}
\endgroup
Here, $\inffunc$ is the information gap function in~\eqref{eqn:inffunc} and
$\tautmo$ is the posterior standard deviation of $g$,
% and $p$ is the dimensionality of $\Zcal$.
and $p,d$ are the dimensionalities of $\Zcal,\Xcal$.
The exponent $q$ depends on the kernel. For the SE kernel, $q=1/(p+d+2)$.
% and $\diamz = \argmax_{z\in\Zcal}\|z-\zhf\|$.
% When $\Zcal=[0,1]^p$, $\diamz \leq \sqrt{p}$.
We filter out the fidelities we consider at time $t$ using three conditions as
specified above.
We elaborate on these conditions in more detail in Section~\ref{sec:fidelSelection}.
If $\candfidelst$ is not empty, we choose the cheapest fidelity in this
set, i.e. $\zt \in \argmin_{z\in\candfidelst}\cost(z)$.
If $\candfidelst$ is empty, we choose $\zt = \zhf$.
% Otherwise, we choose $\zt = \zhf$.

We have summarised the resulting procedure below in Algorithm~\ref{alg:mfgpucbtwo}.
An important advantage of \mfgpucbtwos is that it only requires specifying
the GP hyper-parameters for $\gunc$ such as the kernel $\kernel$.
In practice, this can be achieved by various effective heuristics such as maximising the
GP marginal likelihood or cross validation which are standard in most \bayos{} methods.
In contrast, \mfgpucbones of~\citet{kandasamy16mfbo} requires tuning several other
hyper-parameters.

\insertAlgoMFGPUCBTwo

\subsection{Fidelity Selection Criterion}
\label{sec:fidelSelection}

We will now provide an intuitive justification for the three conditions in the
 selection criterion for $\zt$, i.e., equation~\eqref{eqn:candfidelst}.
Recall that we query $\zhf$ only if $\candfidelst(\xt)$ is empty.
The first condition, $\cost(z) < \cost(\zhf)$ is fairly obvious;
since we wish to optimise $g(\zhf, \cdot)$ and since we are not rewarded for queries at
other fidelities, there
is no reason to consider fidelities that are more expensive than $\zhf$. 

The second condition, $\tautmo(z,\xt) > \gamma(z)$ says that we will only consider
fidelities where the posterior
variance is larger than a threshold
$\gamma(z) = \sqrt{\kernelscale}\inffunc(z)(\cost(z)/\cost(\zhf))^{1/(p+d+2)}$,
 which depends critically on two
quantities, the cost function $\cost$ and the information gap $\inffunc$. As a first step towards parsing this condition, observe that a reasonable multi-fidelity strategy should be inclined to query cheap
fidelities and learn about $\gunc$ before querying expensive fidelities.
Now notice that $\gamma(z)$ is monotonically increasing in $\cost(z)$, therefore, it becomes
easier for a cheap $z$ to satisfy $\tautmo(z,\xt)>\gamma(z)$ and be included in
$\candfidelst$ at time $t$. Moreover, since we choose $\zt$ to be the minimiser of $\cost$ in $\candfidelst$,
a cheaper fidelity will always be chosen over expensive ones if included in $\candfidelst$. 
Second, if a particular fidelity $z$ is far away from $\zhf$, it probably contains less information
about $g(\zhf,\cdot)$. Again, a reasonable multi-fidelity strategy should be discouraged from making such queries. 
This is precisely the role of the information gap $\inffunc$ which is increasing with
$\|z-\zhf\|$.
As $z$ moves away from $\zhf$, $\gamma(z)$ increases and
it becomes harder to satisfy $\tautmo(z,\xt)>\gamma(z)$.
Therefore, such a $z$ is less likely
to be included in $\candfidelst(\xt)$ and be considered for evaluation.
% Our analysis reveals that setting $\gamma$ as in~\eqref{eqn:candfidelst} allows us to
% trade off between the cost and information of the fidelities available to us; 
Our analysis reveals that setting $\gamma$ as in~\eqref{eqn:candfidelst} is a reasonable
trade off between cost and information in the approximations available to us; 
cheaper fidelities cost less, but provide less accurate information about the
function $\func$ we wish to optimise.
% 
% To provide more intuition for the second condition,
% recall that for the SE kernel, $\inffunc(z) \approx \|z-\zhf\|/\hz$ where $\hz$ is the
% bandwidth of $\phiz$.
% If $\gunc$ is very smooth across $\Zcal$, then the lower fidelities provide more
% information about $g(\zhf,\cdot)$.
% One should expect a reasonable strategy
% to exploit such information effectively as it is cheaper.
% Indeed, for fixed $\cost$, large $\hz$ reduces the threshold $\gamma(z)$ across the
% entire $\Zcal$, causing $\candfidelst$ to be nonempty and \mfgpucbtwos to
% consider lower fidelities extensively before proceeding to $\zhf$.
% On the other hand, if $\hz$ is small and $\gunc$ is not smooth across $\Zcal$, the
% threshold $\gamma(z)$ will be large for all $z$. 
% Then $\candfidelst$ vanishes fast and the algorithm will quickly 
% skip lower fidelities and proceed to $\zhf$.
% We demonstrate this behaviour via an example in Section~\ref{sec:experiments}.
%Thus, this behaviour of \mfgpucbtwos is sensible in the SE kernel case. %we are repeating this too often
It is worth noting that the second condition is similar in
spirit to~\citet{kandasamy16mfbo} who proceed from a lower to higher
fidelity only when the lower fidelity variance is smaller than a threshold.
However, while they treat the threshold as a hyper-parameter, we are able to explicitly
specify theoretically motivated values.

% Finally, the third condition in~\eqref{eqn:candfidelst} says we should only
% consider fidelities $z$ satisfying $\inffunc(z) > \inffunc(\diamz)/\betath$. 
% says that we should only pick
% fidelities where the information gap $\inffunc$ is large. 
The third condition in~\eqref{eqn:candfidelst} is
$\inffunc(z) > \inffunc(\diamz)/\betath$. 
Since $\inffunc$ is increasing as we move away from
$\zhf$, it says we should exclude fidelities inside a (small) neighbourhood of $\zhf$.
Recall that if $\candfidelst$ is empty, \boca{} will choose $\zhf$ by default.
But when it is not empty, we want to prevent situations where we get
arbitrarily close to
$\zhf$ but not actually query \emph{at} $\zhf$.
Such pathologies can occur when we are dealing with a continuum of fidelities and
this condition forces \bocas to pick $\zhf$ instead of querying very close to it.
Observe that since $\betat$ is increasing with $t$, this neighborhood is shrinking with time
and therefore the algorithm will eventually have the opportunity to evaluate fidelities
close to $\zhf$.
% Next, we describe our theoretical results.

\subsection{Theoretical Results}
\label{sec:theory}

We now present our main theoretical contributions.
In order to simplify the exposition and convey the gist of our results, we will
only present a simplified version of our theorems.
We will suppress constants, $\polylog$ terms, and other technical
details that arise due to a covering argument in our proofs.
A rigorous treatment is available in Appendix~\ref{sec:analysis}.

\textbf{Maximum Information Gain:}
Up until this point, we have not discussed much about the kernel $\phix$ of the
domain $\Xcal$.
Since we are optimising $\func$ over $\Xcal$, it is natural to expect that this
will appear in the bounds.
\citet{srinivas10gpbandits} showed that the statistical difficulty of GP bandits is
determined by the \emph{Maximum Information Gain} (\mig) which measures the maximum
information a subset of observations have about $\func$. 
We denote it by $\IGn(A)$ where $A$ is a subset of $\Xcal$ and $n$ is the number of
queries to $\func$. We refer the reader to Appendix~\ref{sec:analysis} for a formal definition of \migs. For the current exposition however, it suffices to know that 
$\IGn(A)$ depends
on the domain kernel $\phix$, the number of times $n$ we have queried $\func$, and the
volume $\vol(A)$ of the set $A\subset\Xcal$. 
The latter dependence on $\vol(A)$ will be most important to us.
For instance, when we use an SE kernel for $\phix$, we have
$\IGn(A) \propto \vol(A)\,\logn^{d+1}$%
~\citep{seeger08information}.
~\citet{srinivas10gpbandits} showed that 
the simple regret $S(\capital)$ for \gpucbs after capital $\capital$ can be bounded by,
\begin{align*}
\hspace{-0.1in}
\text{Simple Regret for \gpucb:}\hspace{0.15in}
S(\capital) \lesssim 
  \sqrt{\frac{\IGnc(\Xcal)}{\ncapital}},
\label{eqn:gpucbregret}
\numberthis
\end{align*}
where $\ncapital = \floor{\capital/\cost(\zhf)}$.

In our analysis of \mfgpucbtwos we show that most queries to $\gunc$ at fidelity
$\zhf$ will be confined to a small subset of the domain $\Xcal$ which contains the
 optimum $\xopt$.
More precisely, after capital $\capital$, for any $\alpha\in(0,1)$, we show that 
there exists $\rho>0$ such that the number of queries \emph{outside} the following
set $\Xcalrho$ is less than $\ncapital^\alpha$. 
\begin{align*}
\Xcalrho = \big\{x\in\Xcal\,:\, \funcopt - \func(x)
\leq 2\rho\sqrt{\kernelscale}\,\inffunc(\diamz) \big\}.
\label{eqn:Xcalrho}
\numberthis
\end{align*}
Here, $\inffunc$ is from~\eqref{eqn:inffunc} and $\diamz$ is the $L_2$
diameter of $\Zcal$. %\todo{the following lines are not very clear.}
While it is true that any optimisation algorithm would eventually query extensively
in a neighbourhood around the optimum, a strong result of the above form is
not always possible.
For instance, in the case of \gpucbs, the best achievable bound on the number of queries
in any set that does not contain $\xopt$ is $\ncapital^{1/2}$. 
The fact that the above set $\Xcalrho$ exists relies crucially on the multi-fidelity
assumptions and the fact that our algorithm leverages information from lower
fidelities when querying at $\zhf$.
% and is determined by the smoothness of $\gunc$ across $\Zcal$.
As $\inffunc$ is small when $\gunc$ is smooth across $\Zcal$,  the set $\Xcalrho$ will
be small when the approximations are highly informative about $\gunc(\zhf, \cdot)$. 
To see this more clearly, consider again the case where $\phiz$ is a SE kernel, 
where we have 
 $\Xcalrho \approx \{x\in\Xcal: \funcopt - \func(x) \leq 2\rho\sqrt{p}/\hz\}$.
When $\hz$ is large and $\gunc$ is smooth across $\Zcal$, $\Xcalrho$ is small 
as the right side of the inequality is smaller.
As \mfgpucbtwos confines most of its evaluations to this small set
containing $\xopt$, we will be able to achieve much better regret than \gpucb.
When $\hz$ is small and $\gunc$ is not smooth across $\Zcal$,
the set $\Xcalrho$ becomes large and the advantage of multi-fidelity
optimisation diminishes.

We now provide an informal statement of our main result below.
$\lesssim,\asymp$ will denote inequality and equality ignoring constant and
$\polylog$ terms.

\begin{theorem}[Informal, Regret of \mfgpucbtwo]
\label{thm:regret}
Let $g\sim\GP(0,\kernel)$ where $\kernel$ satisfies~\eqref{eqn:mfkernel}.
Choose $\betat \asymp d\log(t/\delta)$.
Then, for sufficiently large $\capital$ and for all $\alpha \in(0,1)$, there exists $\rho$
depending on $\alpha$ such that the following bound holds w.h.p.
\[
S(\capital) \;\lesssim\;
  \sqrt{\frac{\IGnc(\Xcalrho)}{\ncapital}}
  \;+\;
  \sqrt{\frac{\IGncalpha(\Xcal)}{\ncapital^{2-\alpha}}}
\]
\end{theorem}
In the above bound, the latter term vanishes fast due to the $\ncapital^{-(1-\alpha/2)}$
dependence.
When comparing this with~\eqref{eqn:gpucbregret},
we see that we outperform \gpucbs by a factor of
$\sqrt{\IGnc(\Xcalrho)/\IGnc(\Xcal)} \asymp \sqrt{\vol(\Xcalrho)/\vol(\Xcal)}$
asymptotically.
If $\gunc$ is smooth across the fidelity space, $\Xcalrho$ is small and
the gains over \gpucbs are significant.
If $\gunc$ becomes less smooth across $\Zcal$, the bound decays gracefully, but
we are never worse than \gpucbs up to constant factors.

Theorem~\ref{thm:regret} also has similarities to the bounds 
of~\citet{kandasamy16mfbo} who also demonstrate better regret than \gpucbs by
showing that it is dominated by queries inside a set $\Xcal'$ which contains the optimum.
However, their bounds  depend critically on certain threshold hyper-parameters
which determine the volume of $\Xcal'$ among other  terms in their regret.
The authors of that paper note that their bounds will suffer if these hyper-parameters are not
chosen appropriately, but do not provide theoretically justified methods to make this choice.
In contrast, many of the design choices for \bocas fall out naturally of 
our modeling assumptions.
Beyond this analogue, our results are not comparable to~\citet{kandasamy16mfbo}
as the assumptions are different.

% \subsection{Practical Implementation}
% \label{sec:implementation}

% \subsection{Beyond Continuous Fidelity spaces and Domains}
% \label{sec:extensions}

\textbf{Extensions:}
While we have focused on continuous $\Zcal$ due to their wide ranging
practical applications, many of the ideas here can be extended to other settings.
If $\Zcal$ is a discrete subset of $[0,1]^p$ our work extends straightforwardly.
We reiterate that this will \emph{not} be the same as the finite fidelity
\mfgpucbones algorithm as the assumptions are significantly different.
In particular,~\citet{kandasamy16mfbo} are not able to effectively share
information across fidelities as we do.
We also believe that Algorithm~\ref{alg:mfgpucbtwo} can be extended to arbitrary fidelity
spaces $\Zcal$ given that a kernel can be defined on $\Zcal$.
% Another avenue for future work would be to study different choices for
% $\Zcal$ kernel where 
% These questions are left for future work.
% In particular when $\Xcal$ is a discrete set, the analysis can be extended in a
% straightforward way.
Our results can also be extended to discrete domains $\Xcal$
and various kernels for $\phix$ by adopting techniques 
from~\citet{srinivas10gpbandits}.
As with most nonparametric models, \mfgpucbtwos scales poorly with
dimension due to the dependence on $\IGn$.
For this reason, we also confine our experiments to small $p,d$.
This could be addressed by assuming additional structure on
$\func,\gunc$~\citep{djolonga13highdimbandits,kandasamy15addBO}.

% \input{analysis}

% \vspace{-0.05in}
\insertToyResultsFigure
\vspace{-0.05in}
\vspace{-0.05in}

\section{Experiments}
\label{sec:experiments}

We compare \bocas to the following four baselines: 
(i) \gpucb,
(ii) the \gpeis criterion in \bayo~\citep{jones98expensive}, 
(iii) \mfgpucbones~\citep{kandasamy16mfbo} and
(iv) \mfsko, the multi-fidelity sequential kriging optimisation method
from~\citet{huang06mfKriging}.
All methods are based on GPs and we use the SE kernel for both the fidelity space
and domain.
The first two are not multi-fidelity methods, while the last two are finite multi-fidelity
methods\footnote{%
To our knowledge, the only other work that applies to continuous approximations
is~\citet{klein2015towards} which was developed specifically for hyper-parameter tuning.
Further, their implementation is not made available and is not straightforward to
implement.  }.
We have described the implementation details for all methods in
Appendix~\ref{sec:appImplementation}.

\subsection{Synthetic Experiments}
\label{sec:synthetic}

The results for the first set of synthetic experiments are given in Fig.~\ref{fig:toy}.
The title of each figure states the function used, and the dimensionalities $p,d$ of the
fidelity space and domain.
In all cases, the fidelity space was taken to be $\Zcal=[0,1]^p$ with
$\zhf = \one_p = [1,\dots,1]\in\RR^p$
being the most expensive fidelity.
For \mfgpucbones and \mfsko, we used $3$ fidelities ($2$ approximations) where the
approximations were obtained at $z=0.333\one_p$ and $z=0.667\one_p$ points in the fidelity
space.
To reflect the setting in our theory, we add Gaussian noise to the function value
when observing $g$ at any $(z,x)$.
This makes the problem more challenging than standard global optimisation problems
where function evaluations are not noisy.
The functions $g$, the cost functions $\cost$ and the noise variances $\eta^2$ are given
in Appendix~\ref{sec:appSynthetic}.

The first two figures in Fig.~\ref{fig:toy} are simple sanity checks.
In both cases, $\Zcal=[0,1]$ and $\Xcal=[0,1]$ and the functions were sampled from GPs. 
The GP was made known to all methods, i.e. the methods used the true GP in picking the
next point.
In the first figure, we used an SE kernel with bandwidth $0.1$ for $\phix$ and
$1.0$ for $\phiz$.
The large fidelity bandwidth causes $\gunc$ to be smooth across $\Zcal$ and
\mfgpucbtwos outperforms other baselines in this setting.
The curve starts mid-way in the figure as \bocas is yet to query at $\zhf$ up until
that point.
The second figure uses the same set up as the first except we used an SE
kernel with bandwidth $0.01$ for $\phiz$.
Even though $\gunc$ is highly  unsmooth across $\Zcal$,
\mfgpucbtwos does not perform poorly.
% It immediately skips the lower fidelities (curve starts early) as the threshold
% $\gamma(z)$ is large in~\eqref{eqn:candfidelst} and achieves as good regret
% as \gpucbs and \gpei.
This corroborates a claim that we made earlier that \bocas can naturally adapt
to the smoothness of the approximations.
The other multi-fidelity methods seem to suffer in this setting.

In the remaining experiments, we use some standard benchmarks for global optimisation.
We modify them to obtain $g$ and add noise to the observations.
As the kernel and other GP hyper-parameters are unknown, we learn them by maximising
the marginal likelihood every $25$ iterations.
This is a common heuristic used in the \bayo{} literature.
We outperform all methods on all problems
except in the case of the Borehole function where \mfgpucbones does better.
The last synthetic experiment is the Branin function given in Fig.~\ref{fig:branin}.
We used the same set up as above, but use 10 fidelities for \mfgpucbones and \mfskos
where the $k$\ssth fidelity is obtained at $z = \frac{k}{10}\one_p$ in the fidelity space.
Notice that the performance of finite fidelity methods deteriorate.
In particular, as \mfgpucbones does not share information across fidelities, the
approximations need to be designed carefully for the algorithm to work well.
Our more natural modelling assumptions prevent such pitfalls.
We next present two real examples in astrophysics and hyper-parameter tuning.
We do \emph{not} add noise to the observations, but treat it as optimisation tasks,
where the goal is to maximise the function.

\subsection{Astrophysical Maximum Likelihood Inference}

\label{sec:sn}

We use data on TypeIa supernova for maximum likelihood
inference on $3$ cosmological parameters, the Hubble constant $H_0\in(60,80)$, the
dark matter fraction $\Omega_M\in(0,1)$ and dark energy fraction $\Omega_\Lambda\in(0,1)$,
hence $d=3$.
The likelihood is given by the Robertson-Walker metric, the computation of which
requires a one dimensional numerical integration for each point in the dataset.
Unlike typical maximum likelihood problems, here the likelihood is only accessible
via point evaluations.
\insertRealResultsFigure
We use the dataset from~\citet{davis07supernovae} which has data on $192$ supernovae.
We construct a $p=2$ dimensional multi-fidelity problem where we can choose 
between data set size $N\in[50,192]$ and perform the integration on grids of size 
$G\in[10^2, 10^6]$ via the trapezoidal rule.
As the cost function for fidelity selection, we used $\cost(N,G) = NG$ as
the computation time is linear in both parameters.
Our goal is to maximise the average log likelihood at $\zhf=[192, 10^6]$.
For the finite fidelity methods we use three fidelities with the approximations available
at $z=[97,2.15\times 10^3]$ and $z=[145, 4.64\times 10^4]$ (which correspond to
$0.333\one_p$ and $0.667\one_p$ after rescaling as in Section~\ref{sec:synthetic}).
The results are given in Fig.~\ref{fig:sn} where we plot the maximum average log
likelihood against wall clock time as that is the cost in this experiment.
The plot includes the time taken by each method to tune the GPs and determine the next
points/fidelities for evaluation.
% \bocas outperforms other methods on this problem.

\subsection{Support Vector Classification with $20$ news groups}
\label{sec:news}

We use the $20$ news groups dataset~\cite{joachims1996probabilistic} in a text
classification task.
We obtain the bag of words representation for each document, convert them to tf-idf
features and feed them to a support vector classifier.
The goal is to tune the regularisation penalty and the temperature of the rbf kernel
both in the range $[10^{-2}, 10^3]$. Hence $d=2$.
The support vector implementation was taken from scikit-learn.
We set this up as a $2$ dimensional multi-fidelity problem where we can choose
a dataset size $N\in[5000, 15000]$ and the number of training iterations
$T\in[20, 100]$.
Each evaluation takes the given dataset of size $N$ and splits it up into $5$ to perform
$5$-fold cross validation.
As the cost function for fidelity selection, we used $\cost(N,T) = NT$ as
the training/validation complexity is linear in both parameters.
Our goal is to maximise the cross validation accuracy at $\zhf = [15000, 100]$.
For the finite fidelity methods we use three fidelities with the approximations available
at $z=[8333,47]$ and $z=[11667, 73]$.
The results are given in Fig.~\ref{fig:news} where we plot the average cross validation
accuracy against wall clock time.

% Our goal is to use tune the 

\vspace{-0.05in}
\section{Conclusion}
\label{sec:conclusion}
\vspace{-0.05in}

We studied Bayesian optimisation with continuous approximations, by treating the
approximations as arising out of a continuous fidelity space.
While previous multi-fidelity literature has predominantly focused on a finite number of
approximations, \bocas applies to continuous fidelity spaces and can potentially
be extended to arbitrary spaces.
We bound the simple regret for \bocas and demonstrate that it is better than methods such
as \gpucbs which ignore the approximations and that the gains are determined by the
smoothness of the fidelity space.
When compared to existing multi-fidelity methods, \bocas is able to share information
across fidelities
effectively, has more natural modelling assumptions and has fewer hyper-parameters to tune.
Empirically, we demonstrate that \bocas is competitive with other baselines in synthetic
and real problems.

Going forward, we wish to extend our theoretical results to more general settings.
For instance, we believe a stronger bound on the regret might be possible
if $\phiz$ is a finite dimensional kernel.
Since finite dimensional kernels are typically not radial~\citep{sriperumbudur2016optimal},
our analysis techniques will not carry over straightforwardly.
Another line of work that we have alluded to is to study more general
fidelity spaces with an appropriately defined kernel $\phiz$.

\bibliography{kky,bibGittins}
\bibliographystyle{icml2017}

\onecolumn
\appendix
\section*{\Large Appendix}
\vspace{0.1in}

\section{Some Ancillary Material}
\label{sec:ancillary}

\subsection{Review of \gpucb}
\label{sec:gpucbReview}

We present a review of the \gpucbs algorithm of~\citet{srinivas10gpbandits} which we build
on in this work.
Here we will assume $\func\sim\GP(\zero,\kernel)$ where $\kernel:\Xcal^2\rightarrow\RR$ is
a radial kernel defined on the domain $\Xcal$.
The algorithm is given below.

\insertAlgorithmGPBALGO

To present the theoretical results for \gpucb, we begin
by defining the \emph{Maximum Information Gain} (\mig) which characterises the
statistical difficulty of GP bandits.

\begin{definition}(Maximum Information Gain~\citep{srinivas10gpbandits})
Let $f\sim \GP(\zero, \phix)$.
Consider any $A\subset\RR^d$ and
let $A' = \{x_1, \dots, x_n\} \subset A$ be a finite subset.
Let $f_{A'}, \epsilon_{A'}\in\RR^n$ such that $(f_{A'})_i=f(x_i)$ and
$(\epsilon_{A'})_i\sim\Ncal(0,\eta^2)$.
Let $y_{A'} = f_{A'}+\epsilon_{A'}$.
Denote the Shannon Mutual Information by $I$.
The Maximum Information Gain of $A$ is
\[
\IG_n(A) = \max_{A'\subset A, |A'| = n} I(y_{A'}; f_{A'}).
\]
\label{def:infGain}
\end{definition}
\insertpostspacing

Next, we will need the following regularity conditions on the kernel.
It is satisfied for four times differentiable
kernels such as the SE kernel and \matern{} kernel when
$\nu>2$~\citep{ghosal06gpconsistency}.

% \insertprespacing
\begin{assumption} Let $\func\sim\GP(\zero,\kernel)$, where
$\kernel:\Xcal^2 \rightarrow \RR$ is a stationary kernel.
%  $\kernel(\cdot, x)$ is $L$-Lipschitz for all $x$. 
The partial derivatives
of $\func$  satisfies the following condition.
There exist constants $a, b >0$ such that,
\[
\text{for all $J>0$, $\;$and for all $i \in \{1,\dots,d\}$},\quad \PP\left( \sup_{x} 
\Big|\partialfrac{x_i}{\func(x)}\Big| > J \right)
\leq a e^{-(J/b)^2}.
\]
\label{asm:kernelAssumption}
\end{assumption}

The following theorem is a bound on the simple regret $S_n$~\eqref{eqn:SnDefn} for \gpucb.

\begin{theorem}(\cite{srinivas10gpbandits})
\label{thm:gpucb}
Let $\func\sim\GP(\zero,\kernel)$, where $\Xcal=[0,1]^d$,
$\func:\Xcal\rightarrow\RR$ and the kernel $\kernel$ satisfies
Assumption~\ref{asm:kernelAssumption}).
At each query, we have noisy observations
$y = f(x) + \epsilon$ where $\epsilon\sim\Ncal(0,\eta^2)$.
Denote $C_1 = 8/\log(1+\eta^{-2})$.
Pick a failure probability $\delta\in(0,1)$ and run \emph{\gpucbs} with
  $\betat = 2\log\left(\frac{2\pi^2t^2}{3\delta}\right) + 
  2d\log\left(t^2bdr\sqrt{\frac{4ad}{\delta}}\right)$.
The following holds with probability $>1-\delta$,
% \hspace{-0.2in}
\vspace{-0.02in}
\[
% \PP\left( \forall n\geq 1,\; R_n \leq \sqrt{C_1n\betan\IGn(\Xcal)} \,+\, 2 \right)
% \geq 1-\delta
\text{for all $n\geq 1$,}
\hspace{0.4in}
S_n \leq \sqrt{\frac{C_1\betan\IGn(\Xcal)}{n}} \,+\, \frac{\pi^2}{6}.
\]
\end{theorem}

\subsection{Some Technical Results}
\label{sec:technical}

Here we present some technical lemmas we will need for our analysis.

\begin{lemma}[Gaussian Concentration]
\label{lem:gaussConcentration}
Let $Z\sim\Ncal(0,1)$. Then $\;\PP(Z>\epsilon) \leq \frac{1}{2}\exp(-\epsilon^2/2)$.
\end{lemma}
% \insertprespacing

\begin{lemma}[Mutual Information in GP, \cite{srinivas10gpbandits} Lemma 5.3]
\label{lem:IGformula}
Let $\func\sim\GP(\zero,\kernel)$, $f:\Xcal\rightarrow\RR$ and we observe
$y=f(x) + \epsilon$ where $\epsilon\sim\Ncal(0,\eta^2)$. Let $A$ be a finite
subset of $\Xcal$ and $f_A,y_A$ be the function values and observations on this set
respectively. Then the Shannon Mutual Information
$I(y_A;f_A)$ is,
\[
I(y_A; \func_A) = \frac{1}{2} \sum_{t=1}^{n} \log(1 + \eta^{-2}
  \sigma^2_{t-1}(x_t) ).
\]
where $\sigma^2_{t-1}$ is the posterior GP variance after observing the first $t-1$
points. 
\end{lemma}

Our next result is a technical lemma taken from~\citet{kandasamy16mfbo}.
It will be used in controlling
the posterior variance of our $\func$ and $\gunc$ GPs.

\begin{lemma}[Posterior Variance Bound~\citep{kandasamy16mfbo}]
\label{lem:varbound}
Let $\func\sim(\zero,\kernel)$, $\func:\Ucal\rightarrow\RR$ where
$\kernel(u,u') = \kernelscale\phi(\|u-u'\|)$ and
$\phi$ is a radial kernel.
Upon evaluating $f$ at $u$ we observe $y = f(u) + \epsilon$ where
$\epsilon\sim\Ncal(0,\eta^2)$.
Let $u_1\in\Ucal$ and suppose we have $s$ observations at $u_1$ and no observations
elsewhere.
Then the posterior variance $\kernel'$ (see~\eqref{eqn:gpPost}) at all $u\in\Ucal$
satisfies,
\[
\kernel'(u,u) \leq \kernelscale(1-\phi^2(\|u-u_1\|)) + \frac{\eta^2/s}{ 1+
\frac{\eta^2}{\kernelscale s}}.
\]
\end{lemma}
\textbf{Proof:}
The proof is in Section C.0.1 of~\citet{kandasamy16mfbo} who
prove this result as part of a larger proof.
\hfill\BlackBox

\section{Analysis}
\label{sec:analysis}

We will first state a formal version of Theorem~\ref{thm:regret}.
Recall from the main text where we stated that most evaluations at $\zhf$ are inside
the following set $\Xcalrho$.
\begin{align*}
\Xcalrho = \{x\in\Xcal: \funcopt - \func(x) \leq 2\rho\sqrtks\inffunc(p)\}.
\end{align*}
This is not entirely accurate as it hides a dilation that arises due to a covering
argument in our proofs.
Precisely, we will show that after $n$ queries at any fidelity, 
\bocas will use most of the $\zhf$ evaluations 
in $\Xcalrhon$ defined below using $\Xcalrho$.
\begin{align*}
\Xcalrhon = \big\{\,x\in\Xcal: {\rm B}_2\big(x, \sqrt{d}/n^{\alpha/2d}\big)
\cap \Xcalrhon \neq \emptyset \big\}
\numberthis \label{eqn:Xcalrhon}
\end{align*}
Here ${\rm B}_2(x, \epsilon)$ is an $L_2$ ball of radius $\epsilon$ centred at $x$.
$\Xcalrhon$ is a dilation of $\Xcalrho$ by $\sqrt{d}/n^{\alpha/2d}$.
Notice that for all $\alpha > 0$, as $n\rightarrow\infty$,
$\Xcalrhon$ approaches $\Xcalrho$ at a polynomial rate.
We now state our main theorem below.

\vspace{0.1in}

\begin{theorem}
\label{thm:main}
Let $\Zcal=[0,1]^p$ and $\Xcal=[0,1]^d$. Let $\gunc\sim\GP(\zero,\kernel)$
where $\kernel$ is of the form~\eqref{eqn:mfkernel}. Let $\phix$ satisfy 
Assumption~\ref{asm:kernelAssumption} with some constants $a,b>0$.
% and satisfy assumptions \emph{\Gtwo}, \emph{\Gthree}.
% Let $\kernel$ satisfy
% Assumption~\ref{asm:kernelAssumption} with some constants $a,b$.
Pick $\delta\in(0,1)$ and run \emph{\bocas} with
\[
\betat =
2\log\left(\frac{\pi^2t^2}{2\delta}\right) + 4d\log(t) + 
\max\left\{\,0\,,\, 
2d\log\left(brd\log\bigg(\frac{6ad}{\delta}\bigg) \right)\right\}.
\]
Then, for all $\alpha\in(0,1)$ 
there exists $\rho,\capital_0$ such that with probability at least $1-\delta$ we have
for all $\capital \geq \capital_0$,
\begin{align*}
S(\capital)\;&\leq\;
  \sqrt{\frac{2C_1\beta_{2\ncapital} \IGnn{2\ncapital}(\Xcalrhon)}{\ncapital} }
 \;+\;
  \sqrt{\frac{2C_1\beta_{2\ncapital} \IGnn{2\ncapital^\alpha}(\Xcal)}{\ncapital^{2-\alpha}} }
  \;+\; \frac{\pi^2}{6\ncapital}.
\end{align*}
Here $C_1 = 8/\log(1+\eta^2)$ is a constant and $\ncapital=\floor{\capital/\cost(\zhf)}$.
$\rho$ satisfies %depends only on $\alpha$ and satisfies
$\rho > \rho_0 = \max\{2, 1 + \sqrt{(1+2/\alpha)/(1+d)}\}$.
% $\COSTnought$ depends on several problem dependent quantities including
% the volumes of the sets $\Htcalm$ and consequently on $\rho$.
\end{theorem}
In addition to the dilation,
Theorem~\ref{thm:regret} in the main text also suppresses the constants and $\polylog$ terms.
The next three subsections are devoted to proving the above theorem.
In Section~\ref{sec:setup} we describe some discretisations for $\Zcal$ and $\Xcal$ which
we will use in our proofs.
Section~\ref{sec:techLemmas} gives some lemmas we will need and
Section~\ref{sec:mainproof} gives the proof.

% $\Xcalrhon$ arises due to a covering argument in our proofs.

\subsection{Set Up \& Notation}
\label{sec:setup}

\textbf{Notation:}
Let $U\subset\Zcal\times\Xcal$.
$\Tn(U)$ will denote the number of queries by \bocas at points $(z,x)\in U$ within
$n$ time steps.
When $A\subset\Zcal$ and $B\subset\Xcal$, we will overload notation to denote
$\Tn(A,B) = \Tn(A\times B)$.
% denotes the number of queries by \bocas
% at points $(z,x)\in W\times A$ within time $n$.
For $z\in\Zcal$, $\greaterz$  will denote the fidelities which are more expensive than $z$,
i.e. $\greaterz = \{z'\in\Zcal: \cost(z') > \cost(z)\}$.

We will require a fairly delicate set up before we can prove Theorem~\ref{thm:main}.
Let $\alpha >0$.
All sets described in the rest of this subsection are defined with respect to $\alpha$.
First define
\[
\Hcalntilde = \{(z,x)\in\Zcal\times\Xcal: \funcopt - \func(x) <
2\rho\betanh\sqrtks\inffunc(z) \},
\]
where recall from~\eqref{eqn:inffunc}, $\inffunc(z) = \sqrt{1-\phiz^2(\|z-\zhf\|)}$ is the
information gap function.
We next define $\Hcalnprime$ to be an $L_2$ dilation of $\Hcalntilde$ in the $\Xcal$
space, i.e.
\[
\Hcalnprime = 
\{ (z,x) \in \Zcal\times\Xcal:
{\rm B}_2\big(x, \sqrt{d}/n^{\alpha/2d}\big) \cup
\Hcalntilde \neq \emptyset
 \}.
\]
Finally, we define $\Hcaln$ to be the intersection of $\Hcalnprime$ with all fidelities
satisfying the third condition in~\eqref{eqn:candfidelst}.
That is,
\begin{align*}
\Hcaln = \Hcalnprime \cap \Big\{(z,x)\in\Zcal\times\Xcal: \inffunc(z) >
\inffunc(\sqrt{p})/\betanh \Big\}.
\numberthis \label{eqn:Hcaln}
\end{align*}
In our proof we will use the second condition in~\eqref{eqn:candfidelst} to control the
number of queries in $\Hcaln$.

To control the number of queries outside $\Hcaln$ we first introduce 
a $\frac{\sqrt{d}}{2n^{\frac{\alpha}{2d}}}$-covering of
the space $\Xcal$ of size $n^{\alpha/2}$.
If $\Xcal=[0,1]^d$, a sufficient covering would be an equally spaced grid having
$n^{\frac{\alpha}{2d}}$ points per side.
Let $\{\ain\}_{i=1}^{\nalphabt}$ be the points in the covering.
$\Ain\subset\Xcal$ to be the points in $\Xcal$ which are closest to $\ain$ in $\Xcal$.
Therefore $\Fn = \{\Ain\}_{i=1}^{\nalphabt}$ is a  partition of $\Xcal$.
% We now define $\Fcaln$ to be the union of all $\Ain$ which are outside of $\Hcalntilde$.

Now define $\Qt:2^\Xcal\rightarrow 2^\Zcal$ to be the following function
which maps subsets of $\Xcal$ to subsets of $\Zcal$.
\begin{align*}
\Qt(A) = \Big\{z\in\Zcal: \;\forall\,x\in A, \quad
      \funcopt - \func(x) \geq 2\rho\betath\sqrtks\inffunc(z) \Big\}.
\numberthis\label{eqn:Qt}
\end{align*}
That is, $\Qt$ maps $A\subset\Xcal$ to fidelities where the information gap $\inffunc$
is smaller than $(\funcopt - \func(x))/(2\rho\betath)$ for all $x\in A$.
Next we define $\thetat:2^\Xcal\rightarrow \Zcal$, to be the cheapest fidelity
in $\Qt(A)$ for a subset $A\in\Xcal$. 
\begin{align*}
\thetat(A)\,=\, \arginf_{z\in\Qt(A)} \cost(z).
\numberthis\label{eqn:thetat}
\end{align*}
We will see that \bocas will not query inside an $\Ain\in\Fn$ at fidelities larger
than $\thetat(\Ain)$ too many times (see Lemma~\ref{lem:gtrthetatbound}).
That is, $\Tnzzxx{\greaterzz{\thetan(\Ain)}}{\Ain}$ will be small.
We now define $\Fcaln$ as follows,
\begin{align*}
\Fcaln = \bigcup_{\Ain\subset\Xcal\setminus\Xcalrhon} 
\greaterzz{\thetan(\Ain)} \times \Ain.
\numberthis \label{eqn:Fcaln}
\end{align*}
That is, we first choose $\Ain$'s that are completely outside $\Xcalrhon$
and take their cross product with fidelities more expensive than $\thetat(\Ain)$.
By design of the above sets, and using the third condition in~\eqref{eqn:candfidelst}
we can bound the total number of queries as follows,
% $\Hcaln\cup\Fcaln\cup(\{\zhf\}\times\Xcalrhon) 
% = \Zcal\times\Xcal$. Therefore,
\begin{align*}
n = \Tn(\Zcal,\Xcal) \;\leq\;
  \Tn(\{\zhf\},\Xcalrhon)
  + \Tn(\Fcaln) + \Tn(\Hcaln)
\label{eqn:Tnbreakdown}
\end{align*}
We will show that the last two terms on the right hand side are small for \bocas
and consequently, the first term will be large.
But first, we establish a series of technical results which will be useful in
proving theorem~\ref{thm:main}.

\subsection{Some Technical Lemmas}
\label{sec:techLemmas}

The first lemma proves that the UCB $\utilt$ in~\eqref{eqn:mfucb}
upper bounds $\func(\xt)$ on all the domain points $\{\xt\}_{t\geq 1}$ chosen for
evaluation.
\insertprespacing
\begin{lemma}
\label{lem:ucblemma}
Let $\betat > 2\log(\pi^2t^2/2\delta)$. Then, with probability $>1-\delta/3$, we have
\[
\forall\,t\geq 1,\quad
|\func(\xt) - \mutmo(\xt)|\,\leq\,\betath\sigmatmo(\xt).
\]
\end{lemma}
\textbf{Proof:}
This is a straightforward argument using Lemma~\ref{lem:gaussConcentration}
and the union bound.
At $t\geq 1$,
\begingroup
\allowdisplaybreaks
\begin{align*}
\PP\Big(|\func(x)-\mutmo(x)|>\betath\sigmatmo(x)\Big) &=
\EE\left[\EE\left[ |\func(x)-\mutmo(x)|>\betath\sigmatmo(x) 
  \;\Big|\; \Dcal_{t-1}  \right]\right] \\
&=\EE\left[ \PP_{Z\sim\Ncal(0,1)}\left(|Z|>\betath\right) \right] 
\leq\; \exp\Big(\frac{-\betat}{2}\Big)
\;=\; \frac{2\delta}{\pi^2t^2}.
\end{align*}
\endgroup
In the first step we have conditioned w.r.t $\Dcal_{t-1} = \{(z_i,x_i,y_i)\}_{i=1}^{t-1}$
which allows us to use Lemma~\ref{lem:gaussConcentration}
as $\func(x)|\Dcal_{t-1} \sim\Ncal(\mutmo(x),\sigmasqtmo(x))$.
The statement follows via a union bound over all 
$t\geq 0$ and the fact that $\sum_{t}t^{-2} = \pi^2/6$.
\hfill\BlackBox

Next we show that the GP sample paths are well behaved and that $\utilt(x)$ upper
bounds  $\func(x)$ on a sufficiently dense subset at each time step.
For this we use the following lemma.

\insertprespacing
\begin{lemma}
\label{lem:smoothnesslemma}
Let $\betat$ be as given in Theorem~\ref{thm:main}.
Then for all $t$, there exists a discretisation $G_t$ of $\Xcal$ of size
$(t^2brd\sqrt{6ad/\delta})^d$ such that the following hold.
\vspace{-0.1in}
\begin{itemize}
\item
Let $[x]$ be the closest point to $x\in\Xcal$ in the discretisation.
With probability $>1-\delta/6$, we have
\[
\forall\;t\geq 1,\quad
\forall\;x\in \Xcal,\quad
|\func(x) - \func([x]_t)| \leq 1/t^2. 
\]
\vspace{-0.25in}
\item
With probability $>1-\delta/3$,
for all $t\geq 1$ and for all $a\in G_t$,
$|\func(a) -\mutmo(a)| \leq \betath\sigmatmo(a)$.
\end{itemize}
\end{lemma}
\insertpostspacing
\textbf{Proof:}
The first part of the proof, which we skip here, uses the regularity condition
for $\phix$ in Assumption~\ref{asm:kernelAssumption} and mimics the argument in Lemmas
5.6, 5.7 of~\citet{srinivas10gpbandits}.
The second part mimics the proof of Lemma~\ref{lem:ucblemma} and uses the fact that
$\betat > 2\log(|G_t|\pi^2t^2/2\delta)$.
\hfill\BlackBox

The discretisation in the above lemma is different to the coverings introduced
in Section~\ref{sec:setup}.
The next lemma is about the information gap function in~\eqref{eqn:inffunc}.

\begin{lemma}
\label{lem:inffuncbound}
Let $g\sim\GP(0,\kernel)$, $g:\Zcal\times\Xcal\rightarrow\RR$ and $\kernel$ is
of the form~\eqref{eqn:mfkernel}.
Suppose we have $s$ observations from $g$.
Let $z\in\Zcal$ and $x\in\Xcal$.
Then $\tautmo(z,x) < \alpha$ implies $\sigmatmo(x) < \alpha + \sqrtks\inffunc(z)$.
\end{lemma}
\textbf{Proof:}
The proof uses the observation that for radial kernels, the maximum difference
between the variances at two points $u_1$ and $u_2$ occurs when all $s$ observations
are at $u_2$ or vice versa.
Now we use $u_1 = (z,x)$ and $u_2 = (\zhf,x)$ and apply Lemma~\ref{lem:varbound}
to obtain $\tausqtmo(\zhf,x) \leq \kernelscale(1-\phiz(\|\zhf-z\|))^2 +
\frac{\eta^2/s}{1 + \frac{\eta^2}{s\kernelscale}}$.
However, As $\tausqtmo(z,x) = \frac{\eta^2/s}{1 + \frac{\eta^2}{s\kernelscale}}$
when all observations are at $(z,x)$ and noting that
$\sigmasqtmo(x) = \tausqtmo(\zhf,x) $, we have
$\sigmasqtmo(\zhf,x) \leq \kernelscale(1-\phiz(\|\zhf-z\|))^2 + \tausqtmo(z,x)$.
Since the above situation characterised the maximum difference between
$\sigmasqtmo(x)$ and $\tausqtmo(z,x)$, this inequality is valid for any general
observation set.
The proof is completed using the elementary inequality $a^2 + b^2 \leq (a+b)^2$ for $a,b>0$.
\hfill\BlackBox

We are now ready to prove Theorem~\ref{thm:main}.
The plan of attack is as follows.
We will analyse \bocas after $n$ time steps and bound the number of plays at fidelities
$z\neq\zhf$ and outside $\Xcalrhon$ at $\zhf$.
Then we will show that for sufficiently large $\capital$, the number of \emph{random}
plays $N$ is bounded by $2\ncapital$ with high probability.
Finally we usee techniques from~\citet{srinivas10gpbandits}, specifically the
maximum information gain, to control the simple regret.
However, unlike them we will obtain a tighter bound as we can control the regret due
to the sets $\Xcalrhon$ and $\Xcal\setminus\Xcalrhon$ separately.

% \vspace{0.1in}
% \subsection{Set Up}

\subsection{\textbf{Proof of Theorem~\ref{thm:main}}}
\label{sec:mainproof}

Let $\alpha >0$ be given.
We invoke the sets $\Xcalrhon, \Hcaln,\Fcaln$ in equations~\eqref{eqn:Xcalrhon},
~\eqref{eqn:Hcaln},~\eqref{eqn:Fcaln}
for the given $\alpha$.
The following lemma establishes that for any $A\subset\Xcal$, we will not query
inside $A$ at fidelities larger than $\thetat(A)$~\eqref{eqn:thetat}
too many times.
The proof is given in Section~\ref{sec:gtrthetatbound}.
\begin{lemma}
\label{lem:gtrthetatbound}
Let $A\subset\Xcal$ which does not contain the optimum.
Let $\rho,\betat$ be as given in Theorem~\ref{thm:main}.
Then for all $u>\max\{3, (2(\rho-\rho_0)\eta)^{-2/3}\}$, we have
\[
\PP\Big(\Tnzzxx{\greaterzz{\thetat(A)}}{A} \,>\, u \Big)
\;\leq\; \frac{\delta}{\pi^2}\frac{1}{u^{1+4/\alpha}}
\]
\end{lemma}
To bound $T(\Fcaln)$, we will apply Lemma~\ref{lem:gtrthetatbound} with $u=n^{\alpha/2}$
on all $\Ain\in \Fn$ satisfying
$\Ain\subset\Xcal\setminus\Xcalrhon$.
Since $\Xcalrho\subset\Xcalrhon$, $\Ain$ does not contain the optimum.
As $\Fcaln$ is the union of such sets~\eqref{eqn:Fcaln}, we have for all
$n$ (larger than a constant),
\begin{align*}
\PP(T(\Fcaln) > n^{\alpha}) \;&\leq\;\;
\PP\Big( \exists \Ain\subset\Xcal\setminus\Xcalrhon, \;\;
\Tnzzxx{\greaterzz{\thetat(\Ain)}}{\Ain} \,>\,  n^{\alpha/2} \Big) \\
  &\leq
  \sum_{\substack{\Ain\in \Fn\\ \Ain\subset\Xcal\setminus\Xcalrhon}} \PP\Big(
    \Tnzzxx{\greaterzz{\thetat(\Ain)}}{\Ain} \,>\, n^{\alpha/2} \Big) 
  \leq |\Fn|\frac{\delta}{\pi^2} \frac{1}{n^{\alpha/2 + 2}} 
    \leq \frac{\delta}{\pi^2}\frac{1}{n^2}
\end{align*}
Now applying the union bound over all $n$, we get
$\PP(\forall\,n\geq 1,\;T(\Fcaln) > n^{\alpha}) \,\leq\, \delta/6$.

Now we will bound the number of plays in $\Hcaln$ using the second condition
in~\eqref{eqn:candfidelst}.
We begin with the following Lemma.
The proof mimics the argument in Lemma 11 of~\citet{kandasamy16mfbo} who prove
 a similar result for GPs defined on just the domain, i.e. $\func\sim\GP(\zero,\kernel)$
where $\func:\Xcal\rightarrow\RR$.

\begin{lemma}
\label{lem:mfvarbound}
Let $A\subset\Zcal\times\Xcal$ and the $L_2$ diameter of $A$ in $\Xcal$ be
$\Dx$ and that in $\Zcal$ be $\Dz$.
Suppose we have $n$ evaluations of $g$ of which $s$ are in $A$. 
Then for any $(z,x)\in A$, the posterior variance $\tau'^2$ satisfies, 
\[
\tau'^2(z,x) \leq \kernelscale(1 - \phiz^2(\Dz) \phix^2(\Dx)) + \frac{\eta^2}{s}.
\]
\end{lemma}
Let $\costratio = \cost_{min}/\cost(\zhf)$ where $\cost_{min} = \min_{z\in\Zcal}\cost(z)$.
% To show that we will not query in a certain region at time $n$, it is sufficient to prove
If the maximum posterior variance in a certain region is smaller than $\gamma(z)$,
then we will not query within that region by the second condition
in~\eqref{eqn:candfidelst}.
Further by the third condition, since we will only query at fidelities satisfying
 $\inffunc(z) > \inffunc(\sqrt{p})/\betanh$,
it is sufficient to show that
the posterior variance is bounded by
$\kernelscale \inffunc(\sqrt{p})^2 \costratio^{2q}/\beta_n$ at time $n$ 
to prove that we will not query again in that region.
% This uses the fact that $\inffunc(z) > \inffunc(\sqrt{p})/\betanh$ by the third condition.
For this we can construct a covering of $\Hcaln$ such that 
$1 - \phiz^2(\Dz) \phix^2(\Dx) < \frac{1}{2}\inffunc(\sqrt{p})^2\costratio^{2q}/\betan$.
For any $A\subset\Zcal\times\Xcal$, the covering number, which we denote $\Omega_n(A)$
of this construction
will typically be polylogarithmic in $n$ (See Remark~\ref{rem:construction} below).
Now if there are 
$\frac{2\betan\eta^2}{\costratio^{2q}\inffunc^2(\sqrt{p})\kernelscale} + 1$
queries inside a ball in this covering,
the posterior variance, by Lemma~\ref{lem:mfvarbound} will be smaller than
$\kernelscale \inffunc(\sqrt{p})^2 \costratio^{2q}/\beta_n$. Therefore, 
we will not query any further inside this ball.
Hence, the total number of queries in $\Hcaln$ is
$\Tn(\Hcaln) \leq C_2 \Omega_n(\Hcaln) \frac{\betan}{\costratio^{2q}}
\leq C_3 \vol(\Hcaln) \frac{\polylog(n)}{\poly(\costratio)}$
for appropriate constants $C_2,C_3$.
(Also see Remark~\ref{rem:qchoice}).

Next, we will argue that the number of queries for sufficiently large $\capital$,
is bounded by $\ncapital/2$ where, recall $\ncapital = \floor{\capital/\cost(\zhf)}$.
This simply follows from the bounds we have for $\Tn(\Fcaln)$ and $\Tn(\Hcaln)$.
\[
\Tn(\Zcal\setminus\{\zhf\},\Xcal)
\leq \Tn(\Fcaln) + \Tn(\Hcaln) \leq n^\alpha + \bigO(\polylog(n)).
\]
Since the right hand side is sub-linear in $n$, we can find $n_0$ such that
for all $n_0$, $n/2$ is larger than the right hand side.
Therefore for all $n\geq n_0$, $\Tn(\{\zhf\},\Xcal) > n/2$.
Since our bounds hold with probability $>1-\delta$ for all $n$ we can invert the
above inequality to bound $N$, the random number of queries after capital $\capital$.
We have $N\leq 2\capital/\cost(\zhf)$. We only need to make sure that $N\geq n_0$ which
can be guaranteed if $\capital > \capital_0 = n_0\cost(\zhf)$.

The final step of the proof is to bound the simple regret after $n$ time steps
in \boca. This uses techniques that are now standard in GP bandit optimisation, so we
only provide an outline. We begin with the following Lemma.
\begin{lemma}
\label{lem:IGBoundLemma}
Assume that we have queried $g$ at $n$ points, $(z_t,x_t)_{t=1}^n$ of
which $s$ points are in $\{\zhf\}\times A$ for any $A\subset\Xcal$.
Let $\sigmatmo$ denote the posterior variance of $\func$ at time $t$,
i.e. after $t-1$ queries.
Then, $\sum_{x_t\in A,\zt=\zhf}\sigmasqtmo(x_t)\leq \frac{2}{\log(1+\eta^{-2})}\IGnn{s}(A)$.
Here $\IGnn{s}(A)$ is the MIG of $\phix$ after $s$ queries to $A$ as given
in Definition~\ref{def:infGain}.
\end{lemma}

We now define the quantity $R_n$ below.
Readers familiar with the GP bandit literature might see that it is similar to the notion
of cumulative regret, but we only consider queries at $\zhf$.
\begin{align*}
R_n = \sum_{\substack{t=1\\ \zt=\zhf}}^{n} \funcopt - \func(\xt) \;\;=\;\;
  \sum_{\substack{\zt=\zhf\\\xt\in\Xcalrhon}}
        \funcopt - \func(\xt)
\quad + \quad
  \sum_{\substack{\zt=\zhf\\\xt\notin\Xcalrhon}}
        \funcopt - \func(\xt).
\numberthis
\label{eqn:Rn}
\end{align*}
% We first work with the first term on the right hand side.
% Using a 
For any $A\subset\Xcal$ we can use Lemmas~\ref{lem:ucblemma},~\ref{lem:smoothnesslemma},
and~\ref{lem:IGBoundLemma} and the Cauchy Schwartz inequality to obtain,
\[
\sum_{\substack{\zt=\zhf\\\xt\in A}}
      \funcopt - \func(\xt)
\leq \sqrt{C_1 \Tn(\zhf,A)\betan \IG_{\Tn(\zhf,A)}(A)} + 
\sum_{\substack{\zt=\zhf\\\xt\in A}} \frac{1}{t^2}.
\]
For the first term in~\eqref{eqn:Rn}, we use the trivial bound
$\Tn(\zhf,\Xcalrhon) \leq n$.
For the second term we use the fact that
$\{\zhf\}\times(\Xcal\setminus\Xcalrhon)\subset\Fcaln$ and hence,
$\Tn(\zhf,\Xcal\setminus\Xcalrhon) \leq \Tn(\Fcaln) \leq n^\alpha$.
% Further noting that $ \Xcal\setminus\Xcalrhon \subset \Xcal$, we have
Noting that $A\subset B\implies \IGn(A)\leq\IGn(B)$, we have
$R_n \leq \sqrt{C_1n\betan\IGn(\Xcalrhon)} + \sqrt{C_1n^\alpha\betan\IGnalpha(\Xcal)} 
+ \pi^2/6$.
Now, using the fact that $N \leq 2\ncapital$ for large enough $N$ we have,
\[
R_N \leq  \sqrt{{2C_1\ncapital\beta_{2\ncapital} \IGnn{2\ncapital}(\Xcalrhon)} }
 \;+\;
  \sqrt{{2^\alpha C_1\ncapital^\alpha\beta_{2\ncapital}
\IGnn{2\ncapital^\alpha}(\Xcal)}} 
  \;+\; \frac{\pi^2}{6}.
\]
The theorem now follows from the fact that $S(\capital) \leq \frac{1}{N}R_N$ by definition
and that $N\geq \ncapital$.
The failure instances arise out of Lemmas~\ref{lem:ucblemma},~\ref{lem:smoothnesslemma} and
the bound on $\Tn(\Fcaln)$, the summation of whose probabilities are bounded by
$\delta$.
\hfill\BlackBox

\vspace{0.2in}

\begin{remark}[Construction of covering for the SE kernel]
\label{rem:construction}
\emph{
We demonstrate that such a construction is always possible using the SE kernel.
Using the inequality $e^{-x} \geq 1 - x$ for $x>0$ we have,
\[
1 - \phix^2(\Dx)\phiz^2(\Dz) <
\frac{\Dx^2}{\hx^2} + 
\frac{\Dz^2}{\hz^2} 
\]
% Using the inequality $e^{-x} \leq 1 - x + x^2/2$ for $x>0$ we have,
% \[
% 1 - \phix^2(\Dx)\phiz^2(\Dz) > 
% \frac{\Dx^2}{\hx^2} + 
% \frac{\Dz^2}{\hz^2} -
% \frac{\Dx^4}{2\hx^4} - 
% \frac{\Dz^4}{2\hz^4}
% \]
where $\Dz,\Dx$ will be the $L_2$ diameters of the balls in the covering.
% We will be letting $\Dz,\Dx$ shrink with $n$, so for sufficiently large $n$ the
% right hand side will be larger than 
% $ \frac{\Dx^2}{2\hx^2} + \frac{\Dz^2}{2\hz^2}$.
Now let $h = \min\{\hz, \hx\}$ and choose 
\[
\Dx = \Dz = \frac{h}{2}\frac{\inffunc(\sqrt{p})}{\betanh} \costratio^{q},
\]
via which we have
$1 - \phiz^2(z) \phix^2(x) < \frac{1}{2}\inffunc(\sqrt{p})^2\costratio^{2q}/\betan$
as stated in the proof.
Noting that $\betan\asymp\logn$, using standard results on covering numbers, we can
show that the size of this covering will be
$\logn^{\frac{d+p}{2}}/\costratio^{q(d+p)}$.
A similar argument is possible for Mat\'ern kernels, but the exponent on $\logn$ will be
worse.
}
\end{remark}

\begin{remark}[Choice of $q$ for SE kernel]
\label{rem:qchoice}
\emph{
From the arguments in our proof and Remark~\ref{rem:construction}, we have that
the number of plays in a set $S\subset(\Zcal\times\Xcal)$ is
$T(S)\leq \vol(S) \log(n)^{\frac{d+p+2}{2}}
\left(\frac{\cost(\zhf)}{\cost_{min}}\right)^{q(p+d+2)}$.
However, we chose to work work $\cost_{min}$ mostly to simplify the proof.
It is not hard to see that for $A\subset\Xcal$ and $B\subset\Zcal$
if $\cost(z) \approx \cost'$ for all $z\in B$, then
$\Tn(B,A)\approx \vol(B\times A) \log(n)^{\frac{d+p+2}{2}}
\left(\frac{\cost(\zhf)}{\cost'}\right)^{q(p+d+2)}$.
As the capital spent in this region is $\cost'\Tn(A,B)$,
by picking $q=1/(p+d+2)$ we ensure that the 
capital expended for a certain $A\subset\Xcal$ at all fidelities is roughly the same,
i.e. for any $A$, the capital density in fidelities $z$ such that $\cost(z) <
\cost(\thetat(A))$ will be roughly the same.
\citet{kandasamy16mfbandit} showed that doing so achieved a nearly minimax optimal strategy
for cumulative regret in $K$-armed bandits.
While it is not clear that this is the best strategy for optimisation under GP
assumptions, it did reasonably well in our experiments. We leave it to
future work to resolve this.
}
\end{remark}

\subsubsection{Proof of Lemma~\ref{lem:gtrthetatbound}}
\label{sec:gtrthetatbound}
For brevity, we will denote $\theta = \thetat(A)$.
We will invoke the discretisation $G_t$ used in Lemma~\ref{lem:smoothnesslemma} via which
we have $\utilt([\xopt]_t) \geq \funcopt - 1/t^2$ for all $t\geq 1$.
Let $b = \argmax_{x\in A} \utilt(x)$ be the maximiser of the upper
confidence bound $\utilt$ in $A$ at time $t$.
Now note that,
$\xt\in A \implies \utilt(b) > \utilt([\xopt]_t) \implies
\utilt(b) > \funcopt - 1/t^2$.
We therefore have,
\begingroup
\allowdisplaybreaks
\begin{align*}
\PP\big(\Tnzzxx{\greaterzz{\theta}}{A} \,>\, u \big)\;
&\leq\; 
% \frac{\delta}{\pi^2}\frac{1}{u^{1+4/\alpha}}
\PP\big( \exists t: u+1\leq t \leq n, \;\;
  \utilt(b) > \funcopt - 1/t^2 \;\;\wedge\;\; \tautmo(b) <\gamma(\theta)) 
\\
&\leq\;
  \sum_{t=u+1}^n
\PP\big( \mutmo(b) - \func(b) >  \funcopt - \func(b) 
  - \betath\sigmatmo(b) - 1/t^2
\;\;\wedge\;\; \tautmo(b) <\gamma(\theta)\big) 
\numberthis\label{eqn:sumprobgtr}
% &\hspace{0.2in}\leq\;
%   \sum_{t=u+1}^n \PP_{Z\sim\Ncal(0,1)}\left(Z > (\rho_0-1)\betath\right)
% \;\leq\;
%   \sum_{t=u+1}^n \frac{1}{2}\exp\left(\frac{(\rho_0-1)^2}{2}\betat\right) \\
% &\hspace{0.2in}\leq\;
%   \frac{1}{2}\left(\frac{\delta}{M\pi^2}\right)^{(\rho_0-1)^2}
%   \sum_{t=u+1}^n t^{-(\rho_0-1)^2(2+2d)} 
% \;\leq\;
% \frac{\delta}{M\pi^2} u^{-(\rho_0-1)^2(2+2d) + 1}
% \;\leq\;
% \frac{\delta}{\pi^2} \frac{1}{u^{1+4/\alpha}}.
\end{align*}
\endgroup
We now note that 
\[
\tautmo(b)<\gamma(\theta)\implies \sigmatmo(b) < \gamma(\theta) +
\sqrtks\inffunc(\theta) \leq 2\sqrtks\inffunc(\theta)
\leq \frac{1}{\betath\rho}(\funcopt - \func(b)).
\]
The first step uses Lemma~\ref{lem:inffuncbound}.
The second step uses the fact that $\gamma(\theta) = \sqrtks\inffunc(\theta)
(\cost(z)/\cost(\zhf))^{1/(p+d+2)} \leq \sqrtks\inffunc(\theta)$ and
the last step uses the definition of $\Qt(A)$ in~\eqref{eqn:Qt} whereby we have
$\funcopt - \func(x) \geq 2\rho\betath\sqrtks\inffunc(\theta)$.
Now we plugging this back into~\eqref{eqn:sumprobgtr}, we can bound each term in
the summation by,
\begin{align*}
&\PP\big( \mutmo(b) - \func(b) > (\rho-1)\betath\sigmatmo(b) - 1/t^2 \big) 
\;\leq\;
  \PP_{Z\sim\Ncal(0,1)}\left(Z > (\rho_0-1)\betath\right) \\
&\hspace{0.2in}\leq\;
  \frac{1}{2}\exp\left(\frac{(\rho_0-1)^2}{2}\betat\right) 
\;\leq\;
  \frac{1}{2}\left(\frac{2\delta}{\pi^2}\right)^{(\rho_0-1)^2}
  t^{-(\rho_0-1)^2(2+2d)} 
\;\leq\;
  \frac{\delta}{\pi^2}   t^{-(\rho_0-1)^2(2+2d)} .
\numberthis\label{eqn:termprobgtr}
% \;\leq\;
% \frac{\delta}{M\pi^2} u^{-(\rho_0-1)^2(2+2d) + 1}
% \;\leq\;
% \frac{\delta}{\pi^2} \frac{1}{u^{1+4/\alpha}}.
\end{align*}
In the first step we have used the following facts,
$t>u\geq\max\{3,\,(2(\rho-\rho_0)\eta)^{-2/3}\},\,$
$\pi^2/2\delta >1$ and
$\sigmatmo(b) > \eta/\sqrt{t}$ to conclude,
\begin{align*}
(\rho-\rho_0)\frac{\eta\sqrt{4\logt}}{\sqrt{t}} > \frac{1}{t^2}
\;&\implies\;
(\rho-\rho_0)\cdot
\sqrt{2\log\left(\frac{\pi^2t^2}{2\delta}\right)}\cdot\frac{\eta}{\sqrt{t}} > 
            \frac{1}{t^2} 
\;\implies\;
(\rho-\rho_0) \betath\sigmatmo(b) > \frac{1}{t^2}.
\end{align*}
The second step of~\eqref{eqn:termprobgtr} uses Lemma~\ref{lem:gaussConcentration},
the third step uses the conditions on $\betath$ as given in theorem~\ref{thm:main}
and the last step uses the fact that $\pi^2/2\delta > 1$.
Now plug~\eqref{eqn:termprobgtr} back into~\eqref{eqn:sumprobgtr}.
The result follows by bounding the sum by an integral and  noting that
$\rho_0>2$ and $\rho_0\geq 1+ \sqrt{(1+2/\alpha)/(1+d)}$.
\hfill\BlackBox

\subsubsection{Proof of Lemma~\ref{lem:IGBoundLemma}}

\newcommand{\xAss}[1]{u_{#1}}
\newcommand{\xAt}{\xAss{t}}
\newcommand{\sigmattmo}{\tilde{\sigma}_{t-1}}
\newcommand{\sigmattmosq}{\tilde{\sigma}^2_{t-1}}
Let $A_s = \{\xAss{1},\xAss{2},\dots,\xAss{s}\}$ be the queries in $\{\zhf\}\times A$
in the order they were queried.
Now, assuming that we have queried $\gunc$ only inside $\{\zhf\}\times A$, denote by
$\sigmattmo(\cdot)$, the posterior standard deviation after $t-1$ such queries. Then,
\begin{align*}
\sum_{t:x_t\in A, \zt=\zhf} \sigmasqtmo(x_t) \leq\;
  \sum_{t=1}^s\sigmattmosq(\xAt) 
  \leq\; \sum_{t=1}^s \eta^2\frac{\sigmattmosq(\xAt)}{\eta^2}
  \leq\; \sum_{t=1}^s
    \frac{\log(1+\eta^{-2}\sigmattmosq(\xAt))}{\log(1+\eta^{-2})} 
  \leq \frac{2}{\log(1+\eta^{-2})} I(y_{A_s}; f_{A_s}).
\end{align*}
Queries outside $\{\zhf\}\times A$ will only decrease the variance of the GP so we can
upper bound
the first sum by the posterior variances of the GP with only the queries in
$\{\zhf\}\times A$.
The third step uses the inequality $u^2/v^2 \leq \log(1+u^2)/\log(1+v^2)$.
The result follows from the fact that $\IGnn{s}(A)$
maximises the mutual information among all subsets of size $s$.
\hfill\BlackBox

\section{Addendum to Experiments}
\label{sec:appExperiments}

\subsection{Implementation Details}
\label{sec:appImplementation}

We describe some of our implementation details below.
\begin{itemize}[leftmargin=0.0in]
\vspace{-0.05in}
\item[]
\textbf{Domain and Fidelity space:}
Given a problem with arbitrary domain $\Xcal$ and $\Zcal$, we mapped them to
$[0,1]^d$ and $[0,1]^p$ by appropriately linear transforming the coordinates.

\item[]
\textbf{Initialisation:}
Following recommendations in~\citet{brochu12bo}
all GP methods were initialised with uniform random
queries with $\capital/10$ capital, where $\capital$ is the total capital used in the
experiment.
For \gpucbs and \gpeis all queries were initialised at $\zhf$ whereas
for the multi-fidelity methods, the fidelities were picked at random from
the available fidelities.

\item[]
\textbf{GP Hyper-parameters:}
Except in the first two experiments of Fig.~\ref{fig:toy},
the GP hyper-parameters were learned after initialisation by maximising
the GP marginal likelihood~\citep{rasmussen06gps} and then updated every 25 iterations.
We use an SE kernel for both $\phix$ and $\phiz$ and
instead of using one bandwidth for the entire fidelity space and domain, we learn
a bandwidth for each dimension separately.
We learn the kernel scale, bandwidths and noise variance using marginal likelihood.
The mean of the GP is set to be the median of the observations.

\vspace{-0.05in}
\item[]
\textbf{Choice of $\betat$:} $\betat$, as specified in Theorem~\ref{thm:main} has unknown
constants and tends to be too conservative in practice~\citep{srinivas10gpbandits}.
Following the recommendations in~\citet{kandasamy15addBO} we  set it to be of
the correct ``order''; precisely, $\betat = 0.5d\log(2\ell t + 1)$. 
Here, $\ell$ is the effective $L_1$ diameter of $\Xcal$ and is computed by scaling
each dimension by the inverse of the bandwidth of the SE kernel for that dimension.

\vspace{-0.05in}
\item[]
\textbf{Maximising $\utilt$:}
We used the \directs algorithm~\citep{jones93direct}.

\item[]
\textbf{Fidelity selection:}
Since we only worked in low dimensional fidelity spaces, the set $\candfidelst$
was constructed in practice by obtaining a finely sampled grid of $\Zcal$ and then
filtering out those which satisfied the $3$ conditions in~\eqref{eqn:candfidelst}.
In the second condition of~\eqref{eqn:candfidelst}, the threshold $\gamma(z)$ can
be multiplied up to a constant factor, i.e $c\gamma(z)$ without affecting our theoretical
results.
In practice, we started with $c=1$ but we updated it every 20 iterations via the following
rule: if the algorithm has queried $\zhf$ more than $75\%$ of the time in the last $20$
iterations, we decrease it to $c/2$ and if it queried less than $25\%$ of the time
we increase it to $2c$.
But the $c$ value is always clipped inbetween $0.1$ and $20$.
In practice we observed that the value for $c$ usually stabilised around
$1$ and $8$ although in some experiments it shot up to $20$.
Changing $c$ this way  resulted in slightly better performance in practice.

% \textbf{Multiple Bandwidths in fidelity selection:}
% In theory we assumed a single bandwidth $\hz$ for the fidelity space,
% but in practice we used a bandwidth for each dimensions.
% However, the algorithm uses the bandwidth primarily via the kernel $\phiz$.
% Therefore, we only need to 

\end{itemize}

\subsection{Description of Synthetic Functions}
\label{sec:appSynthetic}

The following are the synthetic functions used in the paper.
\begin{itemize}[leftmargin=0.0in]

\item[]
\textbf{GP Samples:}
For the GP samples in the first two experiments of Figure~\ref{fig:toy} we used an
SE kernel with bandwidth $0.1$ for $\phix$.
For $\phiz$ we used bandwidths $1$ and $0.01$ for the first and second experiments
respectively.
The function was constructed by obtaining the GP function values on a $50\times 50$ grid
in the two dimensional $\Zcal\times\Xcal$ space and then interpolating for evaluations
in between via bivariate splines.
For both experiments we used $\eta^2 = 0.05$ and the cost function $\cost(z) = 0.2 +
6z^2$.

\item[]
\textbf{Currin exponential function:}
The domain is the two dimensional unit cube $\Xcal = [0,1]^2$ and the fidelity 
was $\Zcal = [0, 1]$ with $\zhf = 1$.
We used $\cost(z) = 0.1 + z^2$, $\eta^2=0.5$ and,
\begin{align*}
\gunc(z, x) &= \left(1-0.1(1-z)\exp\left(\frac{-1}{2x_2}\right)\right)
  \left(\frac{2300x_1^3 + 1900x_1^2 + 2092x_1 + 60}{100x_1^3 + 
  500x_1^2 + 4x_1 + 20}\right).
\end{align*}

\item[]
\textbf{Hartmann functions:}
We used 
$g(z,x) = \sum_{i=1}^4 (\alpha_i - \alpha'_i(z)) \exp\big( -\sum_{j=1}^3 A_{ij}
(x_j-P_{ij})^2\big)$.
Here $A, P$ are given below for the $3$ and $6$ dimensional cases and
$\alpha = [1.0, 1.2, 3.0, 3.2]$.
Then $\alpha'_i$ was set as $\alpha'_i(z) = 0.1(1 - z_i)$ if $i\leq p$ for
$i=1,2,3,4$.
We constructed the $p=4$ and $p=2$ Hartmann functions for the $3$ and $6$ dimensional
cases respectively this way.
When $z = \zhf = \one_p$, this reduces to the usual Hartmann function commonly used
as a benchmark in global optimisation.
% The $M$\ssth fidelity function is
% $\funcM(x) = \sum_{i=1}^4 \alpha_i \exp\big( -\sum_{j=1}^3 A_{ij}
% (x_j-P_{ij})^2\big)$ where $A, P\in\RR^{4\times 3}$ are fixed matrices given below and
% $\alpha = [1.0, 1.2, 3.0, 3.2]$. For the lower fidelities we use the same form except
% change $\alpha$ to $\alpha^{(m)} = \alpha + (M-m)\delta$ where $\delta = [0.01, -0.01,
% -0.1, 0.1]$ and $M=3$. The domain is $\Xcal = [0,1]^3$.

For the $3$ dimensional case we used $\cost(z) = 0.05 + (1 - 0.05)z_1^3z_2^2$,
$\eta^2=0.01$ and,
\[
A = 
\begin{bmatrix}
3 & 10 & 30 \\
0.1 & 10 & 35 \\
3 & 10 & 30 \\
0.1 & 10 & 35 
\end{bmatrix},
\quad
P = 10^{-4} \times
\begin{bmatrix}
3689 & 1170 & 2673 \\
4699 & 4387 & 7470 \\
1091 & 8732 & 5547 \\
381 & 5743 & 8828
\end{bmatrix}.
\]
For the $3$ dimensional case we used $\cost(z) = 0.05 + (1 -
0.05)z_1^3z_2^2z_3^{1.5}z_4^{1}$, $\eta^2=0.05$ and,
\[
A = 
\begin{bmatrix}
10 & 3 & 17 & 3.5 & 1.7 & 8 \\
0.05 & 10 & 17 & 0.1 & 8 & 14 \\
3 & 3.5 & 1.7 & 10 & 17 & 8 \\
17 & 8 & 0.05 & 10 & 0.1 & 14 
\end{bmatrix},
\;\;
P = 10^{-4} \times
\begin{bmatrix}
1312 & 1696 & 5569 &  124 & 8283 & 5886 \\
2329 & 4135 & 8307 & 3736 & 1004 & 9991 \\
2348 & 1451 & 3522 & 2883 & 3047 & 6650 \\
4047 & 8828 & 8732 & 5743 & 1091 &  381 \\
\end{bmatrix}.
\]

\item[]
\textbf{Borehole function:}
This function was taken from~\citep{xiong13highAccuracy}.
We first let,
\begin{align*}
\func_2(x) &= \frac{2\pi x_3(x_4-x_6)}
  {\log(x_2/x_1) \left(1 + \frac{2x_7x_3}{\log(x_2/x_1)x_1^2x_8} +
    \frac{x_3}{x_5} \right)}, \\
\func_1(x) &= \frac{5 x_3(x_4-x_6)}
  {\log(x_2/x_1) \left(1.5 + \frac{2x_7x_3}{\log(x_2/x_1)x_1^2x_8} +
    \frac{x_3}{x_5} \right)}. 
\end{align*}
Then we define $g(z,x) = z\func_2(x) + (1-z)\func_1(x)$.
The domain of the function is $\Xcal = [0.05, 0.15; 100, 50K; 63.07K, 115.6K;$
$990, 1110; 63.1, 116; 700, 820;$ $1120, 1680; 9855, 12045]$
and $\Zcal=[0,1]$ with $\zhf = 1$.
We used $\cost(z) = 0.1 + z^{1.5}$ for the cost function and
$\eta^2=5$ for the noise variance.
% We first linear transform the variables to lie in $[0,1]^8$.

\item[]
\textbf{Branin function:}
We use the following function where $\Xcal=[[-5, 10], [0, 15]]^2$ and $\Zcal=[0,1]^3$.
\[
g(z,x) = a(x_2 - b(z_1)x_1^2 + c(z_2)x_1 - r)^2 + s(1-t(z))cos(x_1) + s,
\]
where $a = 1$, $b(z_1)=5.1/(4\pi^2) - 0.01(1-z_1)$
$c(z_2) = 5/\pi - 0.1(1-z_2)$, $r=6$, $s=10$ and $t(z_3)=1/(8\pi) + 0.05(1-z_3)$.
At $z=\zhf = \one_p$, this becomes the standard Branin function used as a benchmark in
global optimisation.
We used $\cost(z) = 0.05 + z_1^{3}z_2^{2}z_3^{1.5}$ for the cost function and
$\eta^2=0.05$ for the noise variance.

\end{itemize}

% \subsection{Details on the Real Experiments}

\end{document}